\documentclass{article}

\usepackage[preprint,nonatbib]{neurips_2021}

\usepackage[utf8]{inputenc} 
\usepackage[T1]{fontenc}    
\usepackage{hyperref}       
\usepackage{url}            
\usepackage{booktabs}       
\usepackage{amsfonts}       
\usepackage{nicefrac}       
\usepackage{microtype}      
\usepackage{xcolor}         
\usepackage{graphicx}
\usepackage{subcaption}
\usepackage{multirow}
\usepackage{hhline}

\usepackage{textcomp}

\usepackage[binary-units=true]{siunitx}
\DeclareSIUnit\flop{FLOP}
\DeclareSIUnit[per-mode=symbol]\floppersec{\flop\per\second}

\def\<#1>{$\langle$\ignorespaces#1\unskip$\rangle$}

\title{Effect of Pre-Training Scale on Intra- and Inter-Domain Full and Few-Shot Transfer Learning for Natural and Medical X-Ray Chest Images}

\author{%
  Mehdi Cherti, \enspace Jenia Jitsev \\
  Juelich Supercomputing Center, Research Center Juelich  \\
  Helmholtz AI \\
  LAION \\
  Juelich, Germany \\
  \texttt{\{m.cherti, j.jitsev\}@fz-juelich.de} \\
}

\begin{document}

\maketitle

\begin{abstract}
Increasing model, data and compute budget scale in the pre-training has been shown to strongly improve model generalization and transfer learning in vast line of work done in language modeling and natural image recognition. However, most studies on the positive effect of larger scale were done in scope of in-domain setting, with source and target data being in close proximity. To study effect of larger scale for both in-domain and out-of-domain setting when performing full and few-shot transfer, we combine here for the first time large, openly available medical X-Ray chest imaging datasets to reach a scale for medical imaging domain comparable to ImageNet-1k, routinely used for pre-training in natural image domain. We then conduct supervised pre-training, while varying network size and source data scale and domain, being either large natural (ImageNet-1k/21k) or large medical chest X-Ray datasets, and transfer pre-trained models to different natural or medical targets\footnote{Repository: \url{https://github.com/SLAMPAI/large-scale-pretraining-transfer}}. We observe strong improvement due to larger pre-training scale for intra-domain natural-natural and medical-medical transfer. For inter-domain natural-medical transfer, we find improvements due to larger pre-training scale on larger X-Ray targets in full shot regime, while for smaller targets and for few-shot regime the improvement is not visible. Remarkably, large networks pre-trained on very large natural ImageNet-21k are as good or better than networks pre-trained on largest available medical X-Ray data when performing transfer to large X-Ray targets. We conclude that substantially increasing model and generic, medical domain-agnostic natural image source data scale in the pre-training can enable high quality out-of-domain transfer to medical domain specific targets, removing dependency on large medical domain-specific source data often not available in the practice.
\end{abstract}

\section{Introduction}
\label{sect:introduction}

Re-using models obtained by pre-training on available source datasets to improve learning performance on upcoming target datasets is core idea behind transfer learning. It has a long history in machine learning field ~\cite{Pratt1991, Pan2009} and was also employed already at the very early rise of deep neural networks in the vision domain~\cite{Razavian2014, Azizpour2016}. Different architectures like AlexNet~\cite{Krizhevsky2012}, OverFeat~\cite{Sermanet2014}, and VGG~\cite{Simonyan2015} were pre-trained on supervised tasks using ImageNet-1k, a publicly available natural image dataset that contains about 1.4 Million images and 1000 classes~\cite{Deng2009a,Russakovsky2015}. After pre-training, the resulting models were taken as off-the-shelf generic features reservoirs and re-used by re-training, or fine-tuning, on various downstream target datasets and tasks, including classification, object detection, and segmentation~\cite{Razavian2014, Azizpour2016}. Importantly, the transfer approach allowed to improve performance on target datasets when compared to training from scratch with randomly initialized weights~\cite{Zhai2019,Mensink2021}. Further, it enabled to train models of good quality also on comparatively small amounts of data, in contrast to large amounts usually required for learning high-quality models when training a deep neural network from scratch.


Recent line of work on scaling laws in language modeling~\cite{Kaplan2020} and vision demonstrated strong improvement for model's ability to generalize on unseen test data when increasing model, data, and compute budget scale during the training. In language modeling, very large Transformer networks pre-trained on very large text data have also shown very strong transfer performance on a broad range of novel tasks, compared to pre-trained models of smaller scale~\cite{Brown2020}. In the same line, experimental studies on large-scale pre-training and transfer in image domain found evidence that increasing network model and data size during pre-training results in transfer performance benefits~\cite{Kolesnikov2020,ridnik2021imagenet21k}.

The majority of the studies looking at the effect of pre-training scale on transfer deal with the intra-domain scenario scenario, where source and target data are close to each other, often originating from the same domain, being for instance natural images. This raises the question whether the observed positive effect of larger scale will also uphold in the inter-domain transfer scenario when using different types of source and target data that are not so closely related. 

To address this, we conduct a series of large-scale pre-training and transfer experiments where we vary not only ResNet model~\cite{He2016a, Kolesnikov2020} and dataset size during pre-training, but also the domain of the source and the target datasets, being either natural or medical X-Ray chest images,  which allows us to study effect of scale on both intra- and inter-domain transfer. To vary source data scale in natural domain, we take either large ImageNet-1k or much larger ImageNet-21k~\cite{Deng2009a}. To vary pre-training data scale for medical X-Ray domain, we combine here, for the first time, large openly available medical X-Ray chest imaging datasets (CheXpert~\cite{Irvin2019}, MIMIC-CXR~\cite{Johnson2019}, PadChest~\cite{Bustos2020}, NIH Chest X-ray14~\cite{Wang2017}) into supersets, with the largest scale comparable to ImageNet-1k. We then transfer to either natural or X-Ray image datasets as target. For transfer, we also vary the operation in either full or few-shot regime, where only few examples per class are shown to the pre-trained models during fine-tuning on a target dataset. As large-scale pre-training requires heavy computational resources, we make use of a state-of-the art supercomputer (JUWELS Booster~\cite{JUWELSBooster2020}) tailored for distributed training to conduct our experiments.


The results we obtain show that both intra- and inter-domain transfer benefit from larger pre-training scale. They also reveal a differentiated picture suggesting that effect of larger scale on transfer is expressed differently in intra- and inter-domain scenario and for full shot and few-shot regime. Remarkably, and also of high relevance for practice, we observe that large networks pre-trained on very large generic natural ImageNet-21k are as good or better than networks pre-trained on largest available medical domain-specific X-Ray superset data when performing full shot transfer to large X-Ray targets. This indicates that high quality models for domain specific medical X-Ray targets can be obtained by increasing scale of the network and of the generic natural image data in the pre-training, without relying on large amount of domain-specific data that is often not available in practice. In contrast to previous studies that dealt with smaller scales~\cite{Raghu2019}, we conclude that inter-domain transfer from natural to medical images benefits from substantially larger pre-training scales.

\section{Background and related work}
\label{sec:background}

\textbf{Scaling laws for generalization and transfer.} Strong evidence that increasing model and data size for the training may result in steady improvement of generalization comes from language modeling studies systematically looking on the dependency of test error on model, data size, and compute budget used for training~\cite{Brown2020,Kaplan2020,Hernandez2021}. The experiments conducted there set up scaling laws with a power law shape and show consistent further decrease of test error when further increasing model, data size, and compute budget over many orders in magnitude hand in hand. For images, a similar line of work by Henighan et al. shows a decrease of test classification top-1 error when fine-tuning on ImageNet pre-trained generative image models of increasing size~\cite{Henighan2020}. Those works use self-supervised training of autoregressive models, in language modeling performed on text and in image domain on image patches, employing transformer networks as a backbone. Additional backup for this line of work comes from studies that revise the dependency of generalization performance on model, data size, and epoch number during training and report double or multi descent curves for the test error~\cite{Belkin2019,Nakkiran2019,Ascoli2020}. There, keeping on increasing model, data size or training time substantially also shows continuous drop in the test error pointing to generalization improvement, for instance when crossing the interpolation threshold and transiting into the over-parameterized regime by scaling up the model size.

\textbf{Improving transfer by scaling up pre-training.} Also improvement in transfer on downstream datasets and tasks is strongly evident from large-scale language modeling experiments. In the study by Brown et al. ~\cite{Brown2020}, large transformer networks in the order of hundred billions of parameters (GPT-3) pre-trained on large text datasets in the order of billions of sentences were shown to have much stronger transfer performance than smaller GPT network models, measured by the test error on different downstream tasks. The difference in transfer performance between different sized models was especially pronounced in the very low data regime when doing zero-shot or few-shot transfer with only few examples available during fine-tuning. Further systematic study on transfer improvements induced by increasing scale was done by Hernandez et al.~\cite{Hernandez2021}, who examined scaling laws for transfer on language modeling tasks in the low-data regime, being defined as transfer using less than $10\%$ of the available target data. The authors have shown that increasing model size in the pre-training decreases test error on the target data, emphasizing that test error improvement cannot be observed when increasing model size and training directly from scratch on the target without pre-training. It was also pointed out that the degree of proximity between the source and the target dataset plays a role when predicting effect of the scale on the transfer performance, leading to a revised version of the scaling law that took this additional dependency into account. 


In the image domain, the performed studies on transfer improvement due to scale have still far less systematic character. Models and datasets used for training on images are 3-4 orders of magnitude behind those studied in language modeling~\cite{Brown2020}. Recently, number of works were starting to employ datasets like ImageNet-21k~\cite{Deng2009a}, YFCC-100M~\cite{Thomee2016}, JFT-300M~\cite{Sun2017} or JFT-3B~\cite{zhai2021scaling} that are much larger than standard ImageNet-1k to pre-train large network models on large data and observe the effect of scaling up on transfer. The work on Big Transfer by Kolesnikov et al.~\cite{Kolesnikov2020} performed supervised classification based pre-training on ImageNet-1k, ImageNet-21k, and JFT-300M using different sized deep residual networks (ResNets~\cite{He2016a}) to study the performance of pre-trained models on transfer across different target datasets. They found consistent improvement in transfer performance when using larger models and larger data during pre-training. In the same direction, works by~\cite{ridnik2021imagenet21k, zhai2021scaling} pre-trained different sized network models on ImageNet-1k, ImageNet-21k or JFT-3B~\cite{zhai2021scaling} also observing consistent transfer improvement when scaling-up model and data size during pre-training.

Studies mentioned above deal with closely related source and target datasets containing mostly natural images data. In general, various works related to testing transfer performance across different target datasets often employ targets that are rather close to pre-training source data, like studies introducing transfer benchmarking datasets that use targets resembling mostly natural image domain ~\cite{Zhai2019,Triantafillou2020} or domain specific transfer studies that stay within their given domain, for instance in medical imaging~\cite{cohen2020limits}.

Only few studies so far attempt to measure transfer performance between datasets that are further apart, for instance natural and medical images, while systematically varying model and data size during pre-training. Work done by Raghu et al. ~\cite{Raghu2019} has found no significant difference between models pre-trained on ImageNet-1k and models trained from scratch on target datasets containing medical images. However, it was not using datasets larger than standard ImageNet-1k or networks larger than standard ResNet-50 for pre-training. Another work examines transfer on CheXpert while varying network model size during pre-training~\cite{Ke2021}. It does find slight benefit for transfer when pre-training with larger models, however it does not vary source data size in the pre-training, using only standard ImageNet-1k as a source. A recent study by Mustafa et al. builds up on Big Transfer work~\cite{Kolesnikov2020} and compares transfer performance of different sized ResNet network models pre-trained on ImageNet-1k, ImageNet-21k, and JFT-300M on different medical imaging target datasets~\cite{Mustafa2021}. Slight evidence for transfer improvement was observed when using larger model and larger dataset sizes during pre-training, with inconsistencies across conditions and datasets, where in some cases no significant benefit from larger pre-training scale was seen. These works all lack comparisons between models pre-trained on natural images data and models pre-trained on medical imaging data of different scale when measuring transfer performance on medical imaging targets.



\section{Experiments \& Results}
\label{sec:experiments}

In order to test the impact of model and data pre-training scale on transfer performance in full and few-shot regime under different source and target data type constellations in intra- and inter-domain scenarios, we conducted experiments on pre-training different sized ResNet models on supervised classification task using either large natural image datasets ImageNet-1k or ImageNet-21k, or compositions of chest X-Ray medical imaging datasets CheXpert, MIMIC-CXR, PadChest, and NIH Chest X-ray14 (see Tab. \ref{tab:datasets_list} for comprehensive list and further details of datasets). The pre-trained models were then fine-tuned on different target datasets that contain either natural or medical images. In the following, we describe the experimental procedures and outcomes in more detail.

\begin{table*}[t]
	\caption{Datasets used as source for pre-training and targets for transfer. All datasets used in the study are publicly available.}
	\begin{minipage}{\textwidth}
		\begin{center} \small
			\setlength{\tabcolsep}{10pt}
			\renewcommand{\arraystretch}{1}
			\begin{tabular}{lcl}
              Dataset & Usage & Size \\
                \hline \hline
                \textit{Natural Images} \\
                \hline
                {ImageNet-1k}~\cite{Deng2009a}  & Source  &       1.4M images, 1000 classes \\
                {ImageNet-21k}~\cite{Deng2009a} & Source   &       14M images, 21842 classes \\
                {CIFAR-10, 100}~\cite{Krizhevsky2009}  & Target  &      60K images, 10,100 classes \\ 
                {Oxford Flowers-102}~\cite{Nilsback2008} & Target & 8K images, 102 classes \\
                {Oxford-IIIT Pet}~\cite{Parkhi2012} & Target & 7.3K images, 37 classes \\
                \hline \hline
                \textit{X-Ray Chest Imaging} \\
                \hline
                {CheXpert}~\cite{Irvin2019} & Source, Target &       224K radiographs of 65K patients, 14 classes \\ 
                {NIH Chest X-ray14}~\cite{Wang2017} & Source, Target    &       112K radiographs of 32K patients, 14 classes \\ 
                {PadChest}~\cite{Bustos2020}    & Source, Target &       160K radiographs of 67K patients, 19 classes \\ 
                {MIMIC-CXR}~\cite{Johnson2019} & Source, Target    &       377K radiographs of 65K patients, 14 classes \\
                \textbf{Total Source X-Ray images} & Source Superset & \textbf{873K} chest radiographs, 229K patients \\
                \hline
                {Tuberculosis}~\cite{Jaeger2014} & Target & 800 radiographs, 800 patients, 2 classes \\
                {COVIDx}~\cite{Wang2020} & Target & 16K radiographs, 15K patients, 2~/~3 classes \\ 
                \hline
			\end{tabular}
		\end{center}	
	\end{minipage}
	\label{tab:datasets_list}
\end{table*}

\subsection{Large-scale pre-training}
\label{subsect:pretrain}

For pre-training, we largely followed the training procedure and used the network architecture of \cite{Kolesnikov2020}.
More concretely, we pre-trained both ResNet-50x1 and ResNet-152x4 (in following R50x1 and R152x4) from \cite{Kolesnikov2020} on different natural image and medical datasets. Smaller R50x1 has 26M weight parameters, while larger R152x4 has 928M parameters. This substantial difference in size allows us to compare the effect of model scaling in the pre-training on subsequent transfer.
The following describes the training procedure and hyper-parameters used for natural image and medical image domain.

\subsubsection{Natural image domain} 
\label{subsect:pretrain_natural}

For natural images, we pre-trained the two models (R50x1 and R152x4) on ImageNet-1k ($\approx 1.4$ Millions images) and the much larger full ImageNet-21k ($\approx 14$ Millions images). For ImageNet-1k and Imagenet-21k models, we used a standard supervised classification setup with softmax as an output activation and cross entropy as a loss.


We followed the training hyper-parameters of \cite{Kolesnikov2020}, with the difference that we used stochastic gradient descent (SGD) with adaptive gradient clipping (AGC) from \cite{brock2021high}, as we found that it helps both pre-training and transfer.
With AGC, we found that the default base learning rate used in \cite{Kolesnikov2020} made training unstable for the ImageNet-21k experiments, so we reduced it from $0.03$ to $0.01$, but otherwise we used a base learning rate of $0.03$.
The rest of the hyper-parameters were similar, namely we used a momentum of $0.9$, $90$ epochs, $\approx 5000$ warmup iterations, a batch size of $4096$, the linear learning rate rescaling rule of \cite{goyal2017accurate}, and the standard step-wise learning rate schedule for ImageNet~\cite{Kolesnikov2020}. For data augmentation, we used the standard random resized crop data augmentation as in \cite{Kolesnikov2020}. In ImageNet-1k experiments, we additionally used RandAugment~\cite{cubuk2020randaugment} and changed the learning rate schedule from step-wise to  cosine annealing~\cite{loshchilov2016sgdr}, as it improved the pre-training results.
In order to speedup training, we used data parallel training with Horovod~\cite{Sergeev2018}, using $256$ A100 GPUs for R152x4 models and $128$ A100 GPUs for R50x1 models. A pre-training on ImageNet-21k with large R152x4 takes about 81 hours using $256$ GPUs, while with small R50x1 it needs about 13.5 hours to finish using $128$ GPUs on JUWELS Booster supercomputer~\cite{JUWELSBooster2020} (see Suppl. Sec. \ref{appendix:distributed_training} and Suppl. Fig. \ref{fig:scaling_R152} for more details on distributed training) 


\subsubsection{Medical image domain}
\label{subsect:pretrain_medical}


For medical data, we pre-trained the two models (R50x1 and R152x4) on combinations of several medical datasets, which as supersets may contain any of the available datasets: CheXpert, MIMIC-CXR, NIH X-ray14, PadChest. We refer to those combinations as X-Ray supersets in following. The largest source X-Ray superset contains about 873K X-Ray chest radiographs (see also Tab. \ref{tab:datasets_list}). The medical datasets are multi-label, as each image can be associated to several diseases. The datasets are combined by finding intersecting labels (diseases) and using the intersected labels as a target. In order to substantially vary data scale for medical domain, we start with single available X-Ray datasets and progressively add other datasets into X-Ray supersets of successively growing size, which provides us with X-Ray source datasets spanning scales from small ($\approx$ 200k samples) to large ($\approx$ 870k samples) to perform pre-training on. For processing the datasets and extracting the labels from raw data, we used TorchXRayVision~\cite{Cohen2020xrv} from the work of \cite{cohen2020limits}.



We followed the literature on medical datasets~\cite{cohen2020limits} and pre-trained using a multi-label setup where we have independent binary tasks, one for each label (disease), and we used sigmoid as an output activation function and binary cross entropy as a loss for each label.

We used the same hyper-parameters as in the natural image domain, except that the base learning rate was set to 0.01 instead of 0.03, as we found that a learning rate of 0.03 led to more overfitting. We followed ~\cite{cohen2020limits} and used a center crop based on the smallest side, then resized the image to $224 \times 224$.
In order to combat overfitting, we used data augmentation from \cite{cohen2020limits}, which included random translation, random rotation, and random scaling. 
In addition to the data augmentation used in \cite{cohen2020limits}, we also do random horizontal flipping.
In order to speedup training, we used data parallel distributed training with Horovod~\cite{Sergeev2018}, using 64 A100 GPUs in all pre-training setups.

\subsection{Fine-tuning and transfer evaluation}
\label{subsect:finetune}

For fine-tuning, we used the BiT-HyperRule\cite{Kolesnikov2020}, which is a heuristic that selects fine-tuning hyper-parameters (learning rate schedule, resolution, usage of MixUp, and total number of steps) based on training set size and image resolution.
We used a batch size of $128$, and an initial learning rate of $0.001$ on all experiments.
Like in \cite{Kolesnikov2020}, we do not use weight decay.
Like in pre-training, we used stochastic gradient descent (SGD) with adaptive gradient clipping (AGC),
as we found it to improve few-shot results. We used a momentum of $0.9$.
In each experiment, the classification head of the pre-trained model was replaced with a new classification head for the fine-tuning task. We fine-tuned all the layers of the network.
For each experiment, we performed $5$ independent runs with different seeds to have an estimate of the variance of the performance.
We ran each fine-tuning experiment on a single A100 GPU.

As in \cite{Kolesnikov2020}, we consider two kinds of setups, few-shot setups (we used 1 or 5 or  10 or  100 or  500 examples per class) and fine-tuning on the full training set.
We used CIFAR-10, CIFAR-100~\cite{Krizhevsky2009}, Flowers-102~\cite{Nilsback2008}, and Oxford-IIIT Pet~\cite{Parkhi2012} for natural image fine-tuning.
For medical image fine-tuning, we used single-label Tuberculosis~\cite{Jaeger2014} and COVIDx~\cite{Wang2020} as small X-Ray targets ($\approx$ 800 and 16k samples each), and multi-label CheXpert, MIMIC-CXR, NIH or PadChest as larger X-Ray targets (magnitude order of 100k-300k samples, see also Tab. \ref{tab:datasets_list}). In addition, to perform few-shot experiments similar to natural domain, we employ PadChest-cl, single-label dataset derived from PadChest, where we keep only images with exactly one label (one disease). For Flowers-102 and COVIDx, since the datasets are strongly imbalanced, we used oversampling.
We measure either final accuracy or mean AUC on the test sets.

\subsection{Results.}
\label{subsect:results}

The experiments allow us to look on both intra- and inter-domain transfer performance following pre-training when varying model size, source data size and source and target dataset domain. In following, we report the obtained results.

\textbf{Effect of scale on intra-domain transfer.} Results we obtain either for natural-natural or medical-medical full shot transfer (Tab. \ref{table:transfer_comparison_all_in_one}) deliver a clear picture showing transfer improvement across target datasets when increasing pre-training model and data scale. Most consistent is the improvement due to increase of network size, while for data scale there are only few single cases where the increase does not result in improvement (e.g, when using large ResNet-152x4 on Pets for natural-natural and for PadChest for medical-medical scenario; see Supplementary for more detailed results for each transfer experiment scenario).

For few-shot transfer, we observe a differentiated picture. In line with previous work, for natural-natural transfer we obtain strong improvement due to larger scale in the very low data regime of 1- or 5-shot transfer, reaching in some cases $20\%-30\%$ absolute difference in test accuracy in favor of larger scale (as seen for CIFAR-100, Fig. \ref{fig:bar_cifar100}). In contrast, for medical-medical scenario, there is no clear evidence for few-shot transfer improvement due to larger scale (Figs. \ref{fig:bar_padchest}, \ref{fig:line_padchest_medical_fewfullshot}, \ref{fig:line_covidx_medical_fewfullshot}, \ref{fig:line_tuberculosis_medical_fewfullshot}; see also Supplementary for further details). Increasing number of shots and approaching full shot regime, the improvement due to scale becomes more and more visible. The observed variance is larger for few-shot transfer experiments, which may suggest less stable fine-tuning in those cases where model has to adapt to target data based on only very limited number of examples.

\begin{table*}[t!]
\centering

\resizebox{\linewidth}{!}{
\begin{tabular}{|c||cc|cc||cc|cc||}
\hline
Target & \multicolumn{4}{c||}{ResNet-50x1} & \multicolumn{4}{c||}{ResNet-152x4}  \\ 
\hline
                    & S-MED   & L-MED     & 1K-NAT & 21K-NAT & S-MED   & L-MED   & 1K-NAT & 21K-NAT   \\  

CIFAR-10\textsuperscript{(1)}         & \color{gray} 56.07 $\pm$ 0.32 & \color{gray} 63.27 $\pm$ 0.30 & 94.26 $\pm$ 0.05 & \textbf{95.78 $\pm$ 0.09} & \color{gray} 74.26 $\pm$ 0.20 & \color{gray} 78.05 $\pm$ 0.18 &  96.93 $\pm$ 0.05 & \color{red} \textbf{97.82 $\pm$ 0.07} \\
CIFAR-100\textsuperscript{(1)}         & \color{gray} 16.64 $\pm$ 0.21 & \color{gray} 18.71 $\pm$ 0.15 & 75.90 $\pm$ 0.05 & \textbf{82.47 $\pm$ 0.21} & \color{gray} 36.29 $\pm$ 0.29 & \color{gray} 37.94 $\pm$ 0.23 & 83.90 $\pm$ 0.09 & \color{red} \textbf{88.54 $\pm$ 0.14} \\
Flowers-102\textsuperscript{(1)}         & \color{gray} 7.05 $\pm$ 0.59 & \color{gray} 6.96 $\pm$ 1.26 & 74.94 $\pm$ 0.99 & \textbf{98.21 $\pm$ 0.22} & \color{gray} 25.19 $\pm$ 0.78 & \color{gray} 23.91 $\pm$ 0.86 &  89.41 $\pm$ 0.25 & \color{red} \textbf{99.49 $\pm$ 0.08} \\
Pets\textsuperscript{(1)}         & \color{gray} 7.06 $\pm$ 0.46 & \color{gray} 7.88 $\pm$ 0.42 & 85.21 $\pm$ 0.58 & \textbf{87.23 $\pm$ 0.18} & \color{gray} 15.07 $\pm$ 0.18 & \color{gray} 16.78 $\pm$ 0.35 &  \color{red} \textit{93.32 $\pm$ 0.30} & \color{red} \textit{93.21 $\pm$ 0.14} \\
\hline
COVIDx\textsuperscript{(1)}         & 68.50 $\pm$ 0.18 & 76.05 $\pm$ 0.21 & 76.30 $\pm$ 1.30 & \textbf{78.35 $\pm$ 1.63} & 78.65 $\pm$ 0.84 & \color{red} \textbf{83.00 $\pm$ 1.16}&  78.10 $\pm$ 0.95 & 78.90 $\pm$ 0.49 \\
Tuberculosis\textsuperscript{(1)}         & 79.83 $\pm$ 0.45 & 81.65 $\pm$ 0.91 & 79.83 $\pm$ 1.50 & \textbf{83.47 $\pm$ 0.83} & 79.01 $\pm$ 0.45 & \color{red} \textbf{90.91 $\pm$ 0.83}&  81.49 $\pm$ 2.23 & 80.83 $\pm$ 2.51 \\
MIMIC CXR\textsuperscript{(2)}         & 84.17 $\pm$ 0.03 & 86.38 $\pm$ 0.03 & 85.41 $\pm$ 0.10 & \textbf{86.82 $\pm$ 0.10} & 87.63 $\pm$ 0.04 & \color{red} \textbf{88.00 $\pm$ 0.03}&  86.85 $\pm$ 0.06 & 87.79 $\pm$ 0.13 \\
CheXpert\textsuperscript{(2)}   & 82.10 $\pm$ 0.07 & \textit{86.66 $\pm$ 0.05} & 84.83 $\pm$ 0.14 & \textit{86.60 $\pm$ 0.14} & 84.92 $\pm$ 0.07 & \color{red} \textit{87.82 $\pm$ 0.03}&  86.82 $\pm$ 0.06 & \color{red} \textit{87.77 $\pm$ 0.07} \\
PadChest\textsuperscript{(2)}         & 68.06 $\pm$ 0.24 & 68.14 $\pm$ 0.21 & 76.72 $\pm$ 0.27 & \textbf{80.99 $\pm$ 0.22} & 75.91 $\pm$ 0.12 & 75.23 $\pm$ 0.17&  79.59 $\pm$ 0.17 & \color{red} \textbf{83.94 $\pm$ 0.19} \\
PadChest-Cl\textsuperscript{(2)}         & 73.01 $\pm$ 0.13 & 78.33 $\pm$ 0.08 & 80.17 $\pm$ 0.17 & \textbf{82.03 $\pm$ 0.17} & 81.79 $\pm$ 0.07 & 82.68 $\pm$ 0.05 &  82.55 $\pm$ 0.05 & \color{red} \textbf{84.02 $\pm$ 0.24} \\
NIH CXR\textsuperscript{(2)}         & 70.11 $\pm$ 0.15 & 74.21 $\pm$ 0.57 & 75.53 $\pm$ 0.47 & \textbf{81.02 $\pm$ 0.57} & 77.95 $\pm$ 0.13 & 78.95 $\pm$ 0.13&  79.82 $\pm$ 0.38 & \color{red} \textbf{82.80 $\pm$ 0.41} \\

\hline
\end{tabular}
}

\caption{\textbf{Varying model and data pre-training scale for intra- and inter-domain transfer.} Pre-training is performed with either natural or medical source data, ordered by increasing scale - being one of large X-Ray datasets (S-MED), larger compositional X-Ray supersets (L-MED, see text and also Tab. \ref{table:transfer_medicalsource_multipletargets} on composition details), ImageNet-1k (1K-NAT),  ImageNet-21k (21K-NAT), using either small ResNet-50x1 or large ResNet-152x4. (1) - Top-1 Acc [$\%$] metric; (2) - mean AUC metric. \textbf{Bold} indicates best transfer performance for a fixed network size and pinpoints the effect of data scale on transfer. \textit{Italics} indicates transfer performance with no significant difference between data scale. \textcolor{red}{Red} indicates best overall performance for a given target. \textcolor{gray}{Gray} indicates medical-natural transfer control experiment.}

\label{table:transfer_comparison_all_in_one}
\end{table*}


\textbf{Effect of scale on inter-domain transfer.} Here we transfer on either small or large medical X-Ray chest imaging targets after pre-training on natural sources of different size, ImageNet-1k (1.4M samples) or much larger ImageNet-21k (14M samples). For all large X-Ray targets (NIH, CheXpert, PadChest, PadChest-cl or MIMIC-CXR) we observe clear full-shot transfer improvement due to larger pre-training scale (Tab. \ref{table:transfer_comparison_all_in_one}). The effect is consistent for both model and data scale across large X-Ray targets.

For small X-Ray targets (Tuberculosis and COVIDx), we do not observe such consistent improvement due to larger scale. For instance, while we see improvement due to larger data scale for small ResNet-50x1 on both small targets, the improvement is not there when increasing network size. There is also no evidence for systematic positive effect of larger scale on few-shot transfer, neither for large nor for small X-Ray targets (Figs. \ref{fig:bar_padchest}, \ref{fig:line_padchest_natural_fewfullshot}, \ref{fig:line_covidx_natural_fewfullshot}, \ref{fig:line_tuberculosis_natural_fewfullshot}). Again, variance observed in few-shot regime is large, and is getting smaller and smaller when increasing number of shots and moving towards full shot transfer.

\begin{table*}[t!]
\centering
\begin{scriptsize}
\begin{tabular}{|c||ccc||ccc||}
\hline
\multirow{2}{*}{Target} & \multicolumn{3}{c||}{ResNet-50x1} & \multicolumn{3}{c||}{ResNet-152x4}  \\ 
                                      & CheXpert   & +PadChest  & +MIMIC  & CheXpert  & +PadChest  & +MIMIC  \\  
\hline
NIH      & 70.11 $\pm$ 0.15 & \textit{73.37 $\pm$ 0.38} & \textit{74.21 $\pm$ 0.57} & 77.95 $\pm$ 0.13 & 78.16 $\pm$ 0.13 & \color{red} \textbf{78.95 $\pm$ 0.13}  \\ 
\hline
                                      & CheXpert   & +MIMIC  & +NIH  & CheXpert  & +MIMIC  & +NIH  \\  
PadChest       & 68.06 $\pm$ 0.24 & \textbf{70.07 $\pm$ 0.49} & 68.14 $\pm$ 0.21 & \color{red} \textit{75.91 $\pm$ 0.12} & \color{red} \textit{75.81 $\pm$ 0.07} & 75.23 $\pm$ 0.17  \\ 
PadChest-Cl       & 73.01 $\pm$ 0.13 & \textit{78.44 $\pm$ 0.04} & \textit{78.33 $\pm$ 0.08} & 81.79 $\pm$ 0.07 & \color{red} \textbf{83.14 $\pm$ 0.04} & 82.68 $\pm$ 0.05  \\
\hline
                                    & PadChest   & +MIMIC  & +NIH  & PadChest  & +MIMIC  & +NIH  \\  
CheXpert       & 82.10 $\pm$ 0.07 & \textit{86.56 $\pm$ 0.08} & \textit{86.66 $\pm$ 0.05} & 84.92 $\pm$ 0.07 & \color{red} \textbf{88.03 $\pm$ 0.03} & 87.82 $\pm$ 0.03  \\
\hline
                                      & CheXpert   & +PadChest  & +NIH  & CheXpert  & +PadChest  & +NIH  \\  
MIMIC CXR       & 84.17 $\pm$ 0.03 & 86.19 $\pm$ 0.03 & \textbf{86.38 $\pm$ 0.03} & 87.63 $\pm$ 0.04 & \color{red} \textbf{88.13 $\pm$ 0.03} & 88.00 $\pm$ 0.03  \\
\hline
\end{tabular}

\vspace*{2mm}

\begin{tabular}{|c||cccc||cccc||}
\hline
\multirow{2}{*}{Target} & \multicolumn{4}{c||}{ResNet-50x1} & \multicolumn{4}{c||}{ResNet-152x4}  \\ 
                                      & CheXpert   & +MIMIC  & + NIH & +PadChest & CheXpert  & +MIMIC  & +NIH & +PadChest      \\  
\hline
COVIDx\textsuperscript{(1)}         & 68.50 $\pm$ 0.18 & 75.10 $\pm$ 1.52 & 75.60 $\pm$ 0.45 & \textbf{76.05 $\pm$ 0.21} & 78.65 $\pm$ 0.84 & 81.65 $\pm$ 0.74 & 80.80 $\pm$ 1.10 & \color{red} \textbf{83.00 $\pm$ 1.16}  \\
Tuberculosis\textsuperscript{(1)}   & 79.83 $\pm$ 0.45 & 78.84 $\pm$ 1.25 & \textit{81.32 $\pm$ 0.74} & \textit{81.65 $\pm$ 0.91} & 79.01 $\pm$ 0.45 & 84.63 $\pm$ 0.74 & 87.93 $\pm$ 0.74 & \color{red} \textbf{90.91 $\pm$ 0.83}  \\
\hline
\end{tabular}

\end{scriptsize}
\caption{\textbf{Intra-domain transfer using different sized medical X-Ray source data for pre-training with different sized ResNets, using different X-Ray targets} (Mean AUC metric; \textbf{above}: larger, \textbf{below}: smaller targets). "+" indicates addition of a set into a successively larger source superset. For each target, source datasets correspond to either S-MED datasets or L-MED supersets reported in Tab. \ref{table:transfer_comparison_all_in_one}.  Clear transfer improvement is evident for increasing both model and data scale. While improvement via scaling up the model is consistent, scaling up pre-training data shows in some cases no significant change, especially when adding a smaller dataset like NIH which leads to a rather small increase in data scale. \textbf{Bold}, \textit{Italics}, \textcolor{red}{Red} and \textcolor{gray}{Gray} indicate the same as in Tab. \ref{table:transfer_comparison_all_in_one}}
\label{table:transfer_medicalsource_multipletargets}
\end{table*}

Remarkably, when further comparing intra- and inter-domain transfer performance, we observe that large ResNet-152x4 pre-trained on very large generic natural ImageNet-21k are as good or better than networks pre-trained on largest available medical domain specific X-Ray superset data when performing full shot transfer to large X-Ray targets (Tab. \ref{table:transfer_comparison_all_in_one}, Figs. \ref{fig:line_padchest_natural_fewfullshot},  \ref{fig:line_padchest_medical_fewfullshot}). This fits into overall picture of larger model and data pre-training scale improving transfer on larger targets observed here, as ImageNet-21k has order of magnitude larger scale than the largest X-Ray superset constructed for this study. 

For control, we also conduct inter-domain transfer in medical-natural scenario (grayed out area in Tab. \ref{table:transfer_comparison_all_in_one}). There, as expected, after pre-training on specific X-Ray data the performance on natural targets is severely diminished when compared to transfer performance shown by models pre-trained on natural images. The effect of scale is though still preserved in this control scenario, clearly favoring large over the small network scale.

\section{Discussion \& Conclusion}
\label{sec:discussion_and_conclusion}

Our observations of the transfer performance dependency on the pre-training model and data scale and on source and target domain alignment suggest that both intra- and inter-domain transfer benefit from larger pre-training scale. The effect of pre-training scale depends however on transfer conditions, revealing a differentiated picture of when larger scale may lead to transfer improvement.



\begin{figure*}[t!]
\begin{subfigure}{.35\textwidth}
  \centering
  \includegraphics[width=\linewidth]{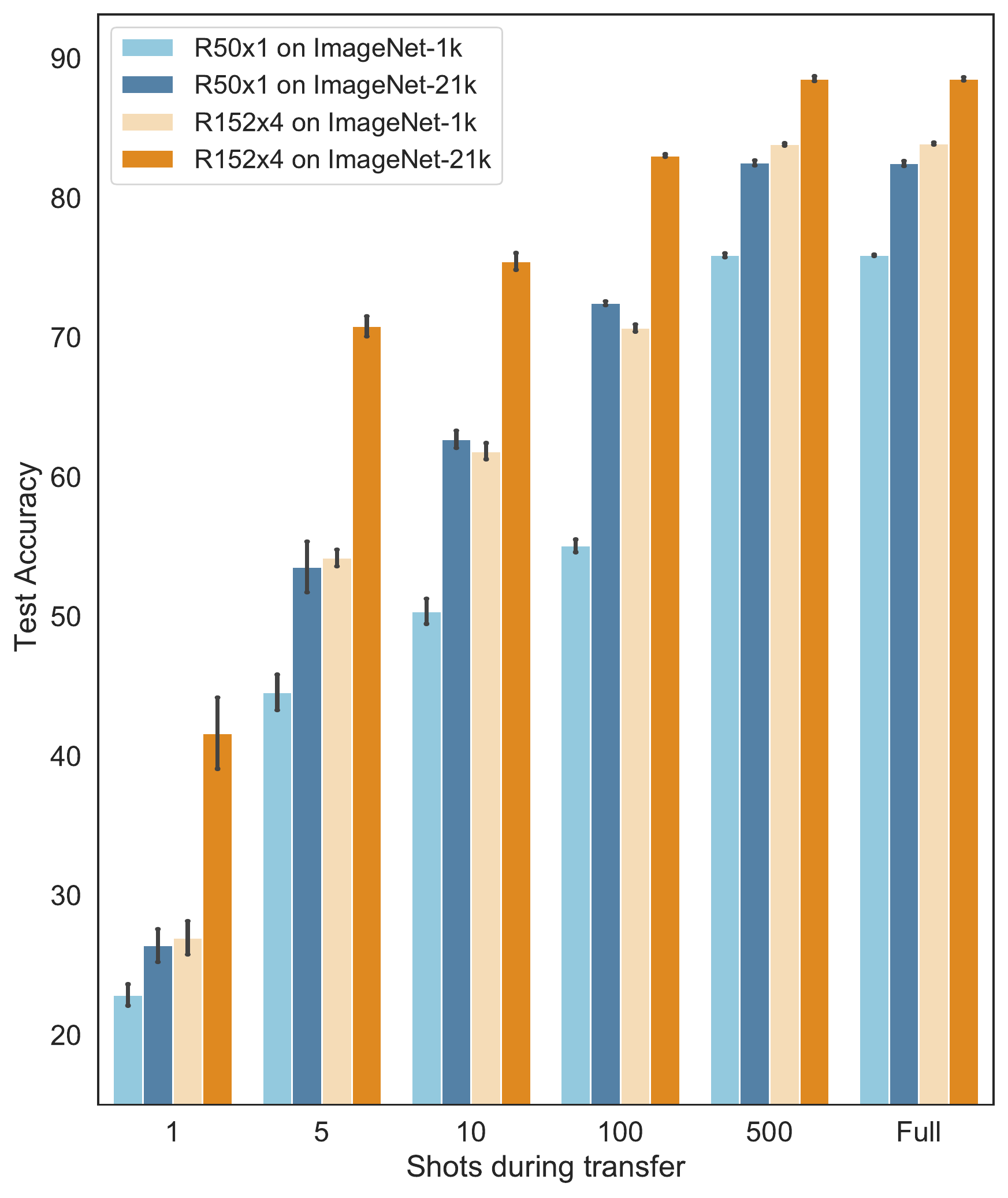}
  \caption{CIFAR-100}
  \label{fig:bar_cifar100}
\end{subfigure}
\begin{subfigure}{.65\textwidth}
  \centering
  \includegraphics[width=\linewidth]{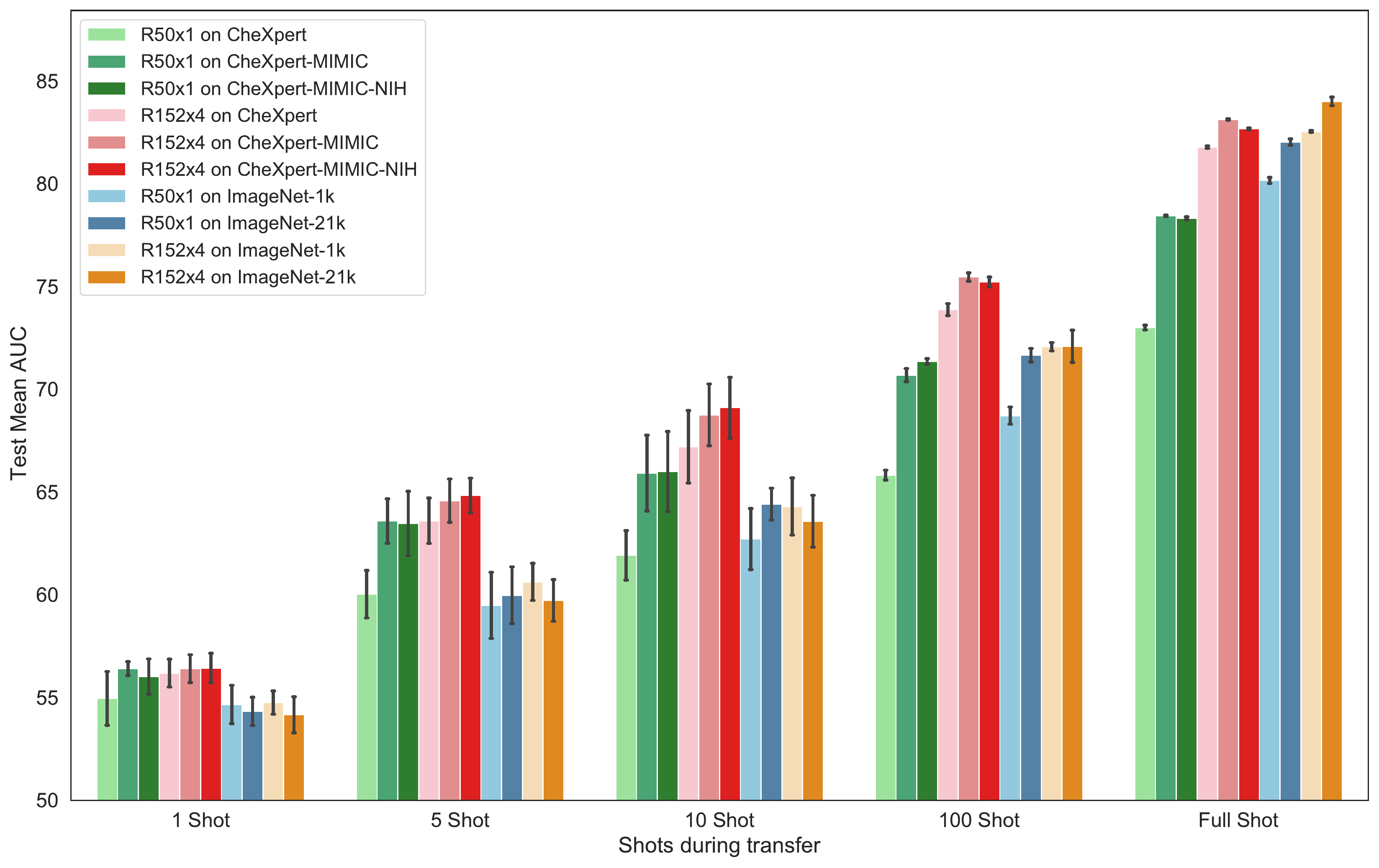}
  \caption{PadChest-Cl}
  \label{fig:bar_padchest}
\end{subfigure}
\caption{Few- and full shot transfer performance on a natural and a medical X-Ray target when varying model and data scale in pre-training.
}
\label{fig:bar_few_full}
\end{figure*}
\begin{figure*}[t!]
\centering
\begin{subfigure}{.4\textwidth}
  \centering
  \includegraphics[width=\linewidth]{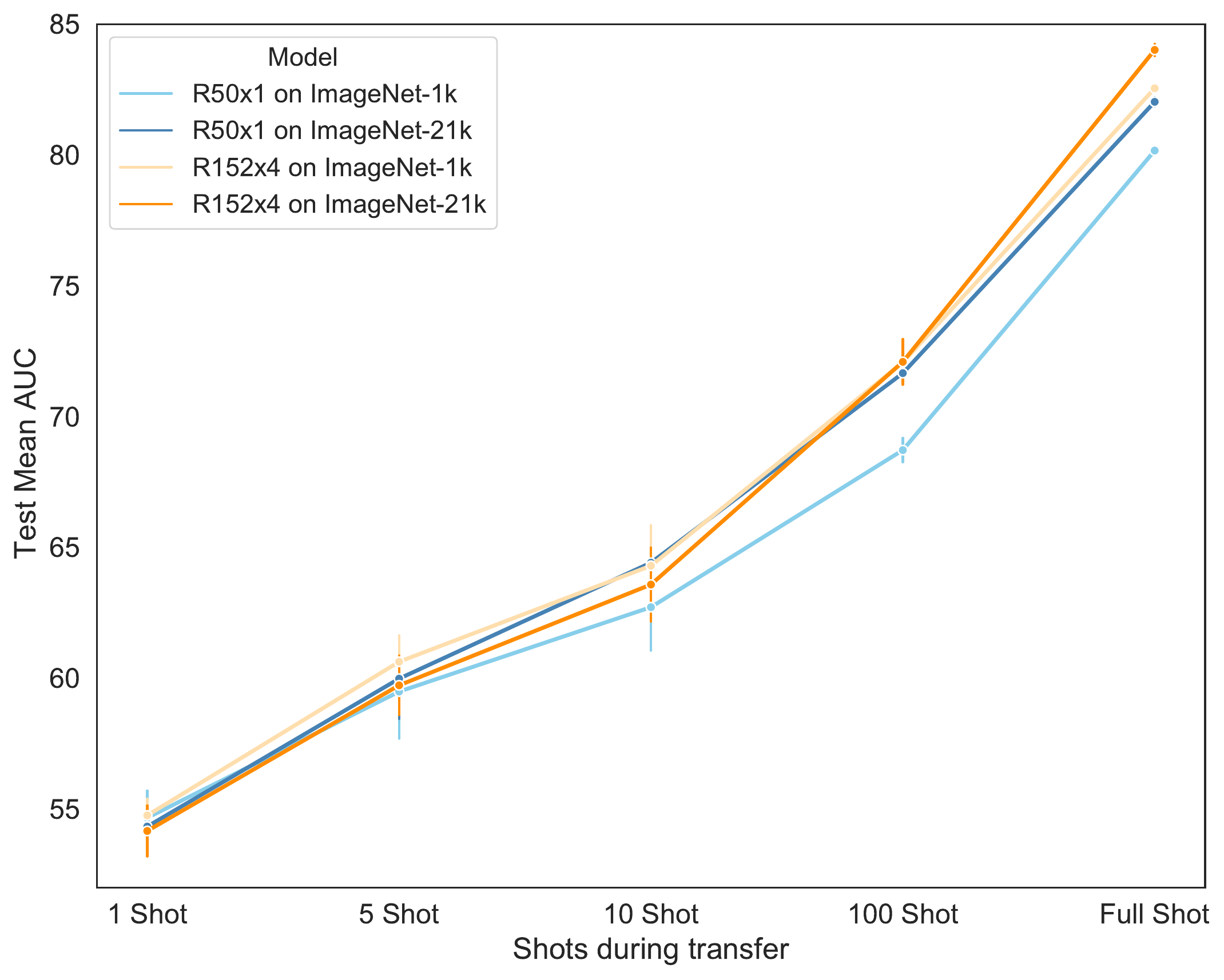}
  \caption{PadChest-Cl, natural source}
  \label{fig:line_padchest_natural_fewfullshot}
\end{subfigure}
\begin{subfigure}{.4\textwidth}
  \centering
  \includegraphics[width=\linewidth]{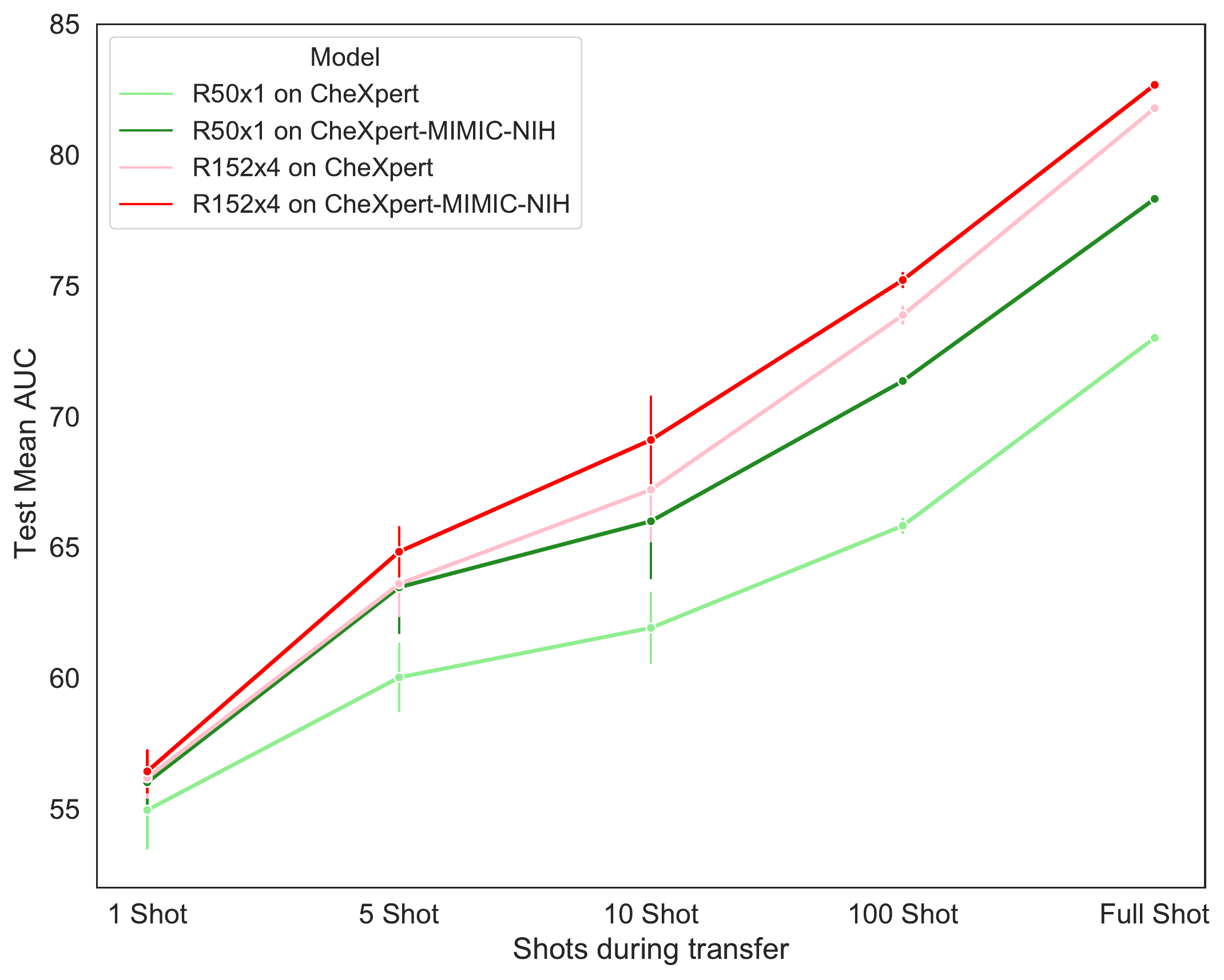}
  \caption{PadChest-Cl, medical source}
  \label{fig:line_padchest_medical_fewfullshot}
\end{subfigure}
\begin{subfigure}{.4\textwidth}
  \centering
  \includegraphics[width=\linewidth]{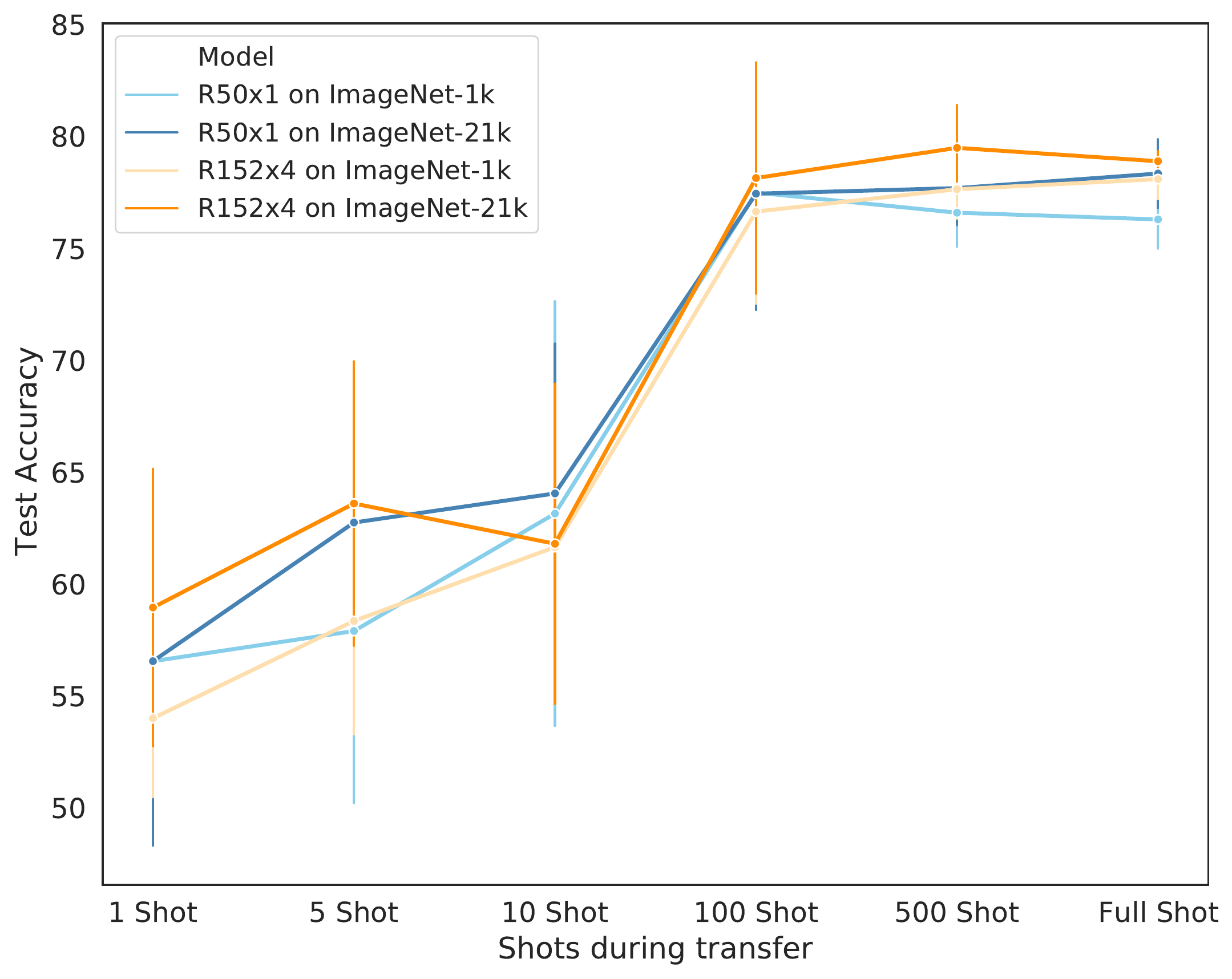}
  \caption{COVIDx, natural source}
  \label{fig:line_covidx_natural_fewfullshot}
\end{subfigure}
\begin{subfigure}{.4\textwidth}
  \centering
  \includegraphics[width=\linewidth]{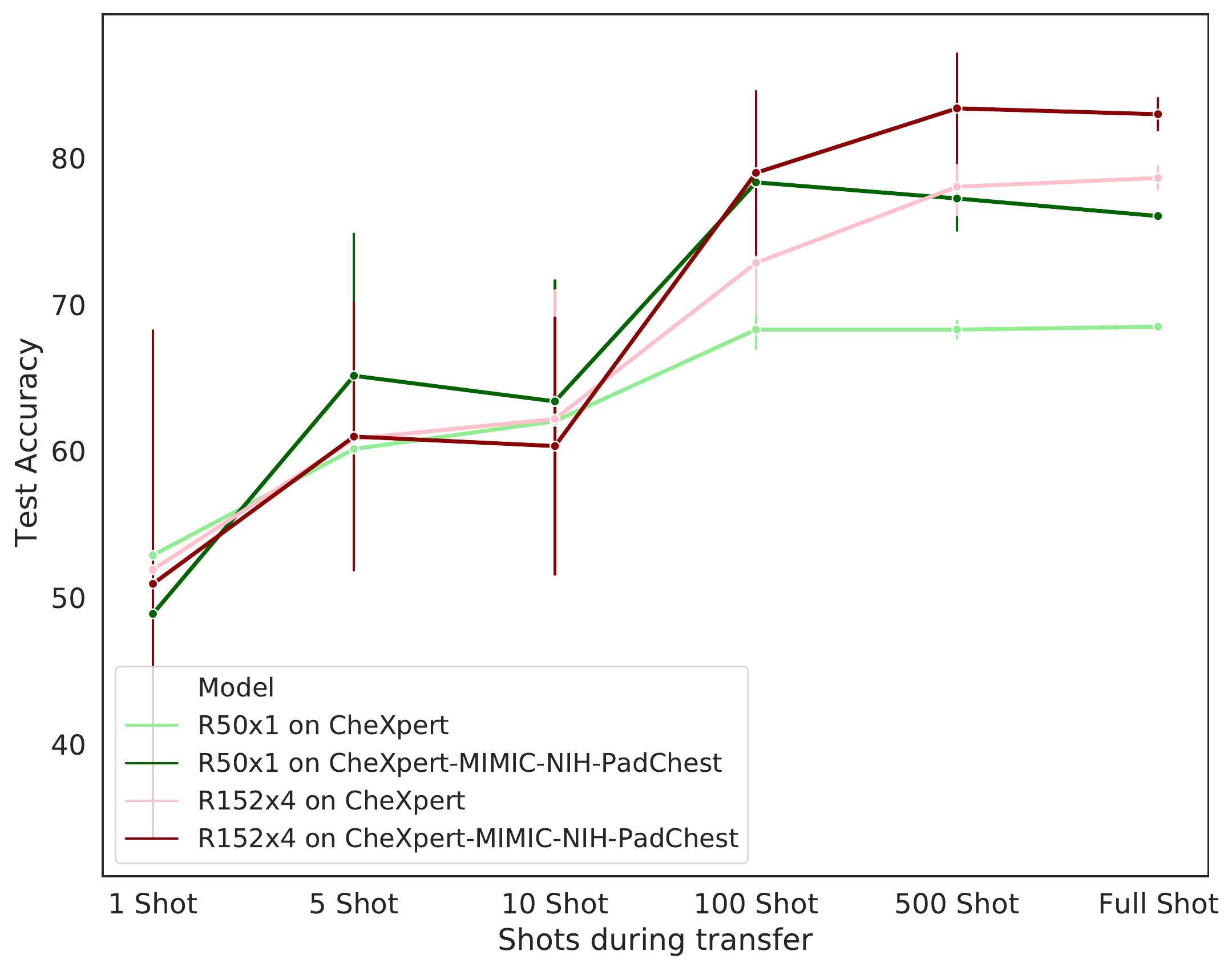}
  \caption{COVIDx, medical source}
  \label{fig:line_covidx_medical_fewfullshot}
\end{subfigure}
\begin{subfigure}{.4\textwidth}
  \centering
  \includegraphics[width=\linewidth]{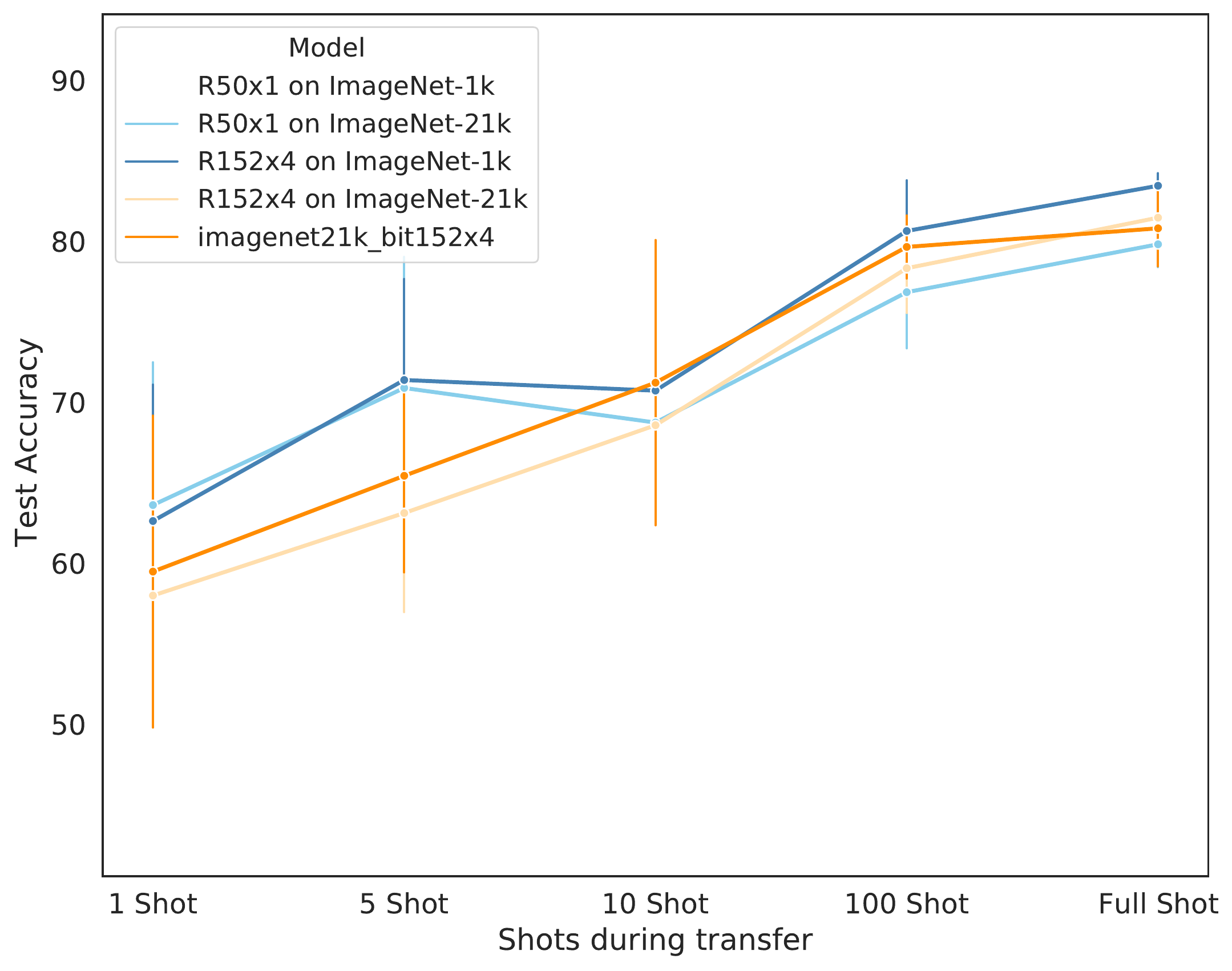}
  \caption{Tuberculosis, natural sources}
  \label{fig:line_tuberculosis_natural_fewfullshot}
\end{subfigure}
\begin{subfigure}{.4\textwidth}
  \centering
  \includegraphics[width=\linewidth]{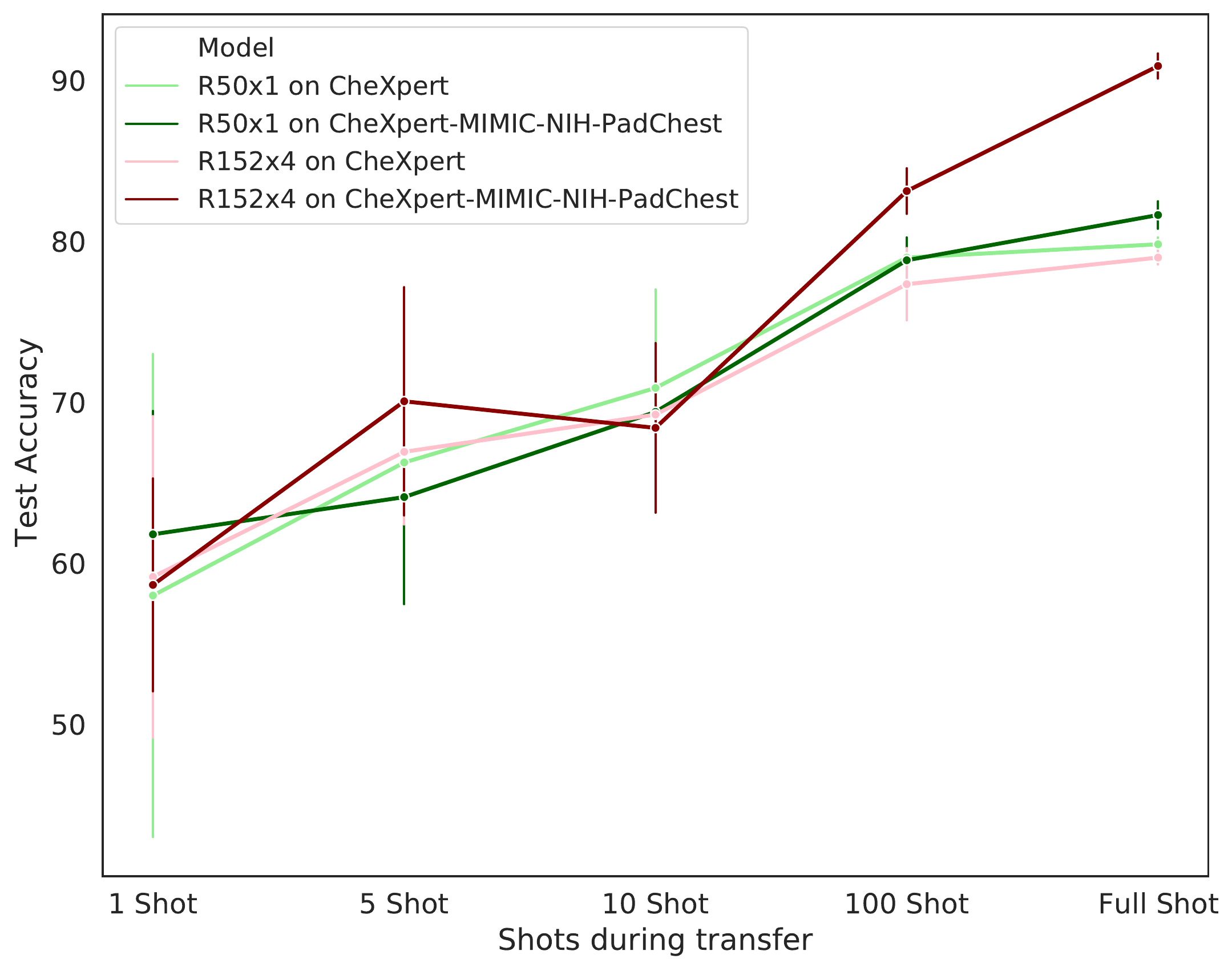}
  \caption{Tuberculosis, medical sources}
  \label{fig:line_tuberculosis_medical_fewfullshot}
\end{subfigure}

\caption{
Few-shot and full shot transfer performance on medical X-Ray targets of different size for intra- and inter-domain scenarios, medical-medical or natural-medical. Each color represents a combination of model and data scale during pre-training. Positive effect of larger scale on full-shot transfer is evident for large X-Ray target PadChest-Cl (top), both for intra- and inter-domain scenario. Remarkably, by increasing natural image data scale the largest network outperforms there the same network pre-trained on large X-Ray data (compare top (a) and (b), full-shot setting). Effect of scale is not evident for few-shot transfer and for small targets in inter-domain setting.}
\label{fig:line_fewfullshot}
\end{figure*}

\textbf{Larger pre-training scale improves intra-domain transfer} We obtain evidence that both natural-natural domain transfer and medical-medical domain transfer are improved when increasing model and data size during pre-training (Tab. \ref{table:transfer_comparison_all_in_one}; see also Suppl. material). For natural-natural transfer scenario, the improvement is evident for both full and few-shot regime. Increasing data size by using ImageNet-21k instead of ImageNet-1k or increasing network model size by using ResNet-152x4 instead of smaller ResNet-50x1 creates strong, consistent boost in transfer performance across all natural target datasets, which is in line with previous observations~\cite{Kolesnikov2020, ridnik2021imagenet21k}. Improvement is especially pronounced in few-shot regime (e.g, Fig. \ref{fig:bar_cifar100}, see also Suppl. material), also adding evidence for more data efficient transfer due to larger scale. The picture is more differentiated for medical-medical transfer scenario. For full transfer regime, the improvement due to larger pre-training scale is clear and consistent across different targets (Tab. \ref{table:transfer_comparison_all_in_one}). In few-shot transfer however, in contrast to natural-natural scenario (Fig. \ref{fig:bar_cifar100}), there are no benefits due to larger scale (Fig. \ref{fig:bar_padchest}, \ref{fig:line_padchest_medical_fewfullshot},  \ref{fig:line_covidx_medical_fewfullshot}, \ref{fig:line_tuberculosis_medical_fewfullshot}). Here we have to keep in mind that both absolute data size and increase in data scale we obtain by going from ImageNet-1k to ImageNet-21k (14M samples) in natural pre-training is much stronger than what we achieve in medical pre-training, going from one of X-Ray datasets to the largest combined X-Ray superset that still has much smaller data volume ($\approx$ 870k samples) than ImageNet-21k. This difference in data scale may also explain the observed differences in few-shot regime, while we also cannot rule out that domain type (natural or medical) could as well play an important role in determining how transfer is affected by pre-training scale.

\textbf{Larger pre-training scale improves inter-domain transfer for larger targets.} In contrast to intra-domain transfer, where natural-natural or medical-medical source and target datasets are closely related, in natural-medical inter-domain transfer the source and target are much further apart. It is therefore not trivial that effect of pre-training scale brings similar or any improvement in this case as well. Strong discrepancy between source and target may render transfer ineffective, as it was indeed observed in previous studies on natural-medical transfer done on smaller scales~\cite{Raghu2019}. In contrast to these studies, we do find here significant positive effect of larger pre-training scale on inter-domain natural-medical full shot transfer for larger medical targets (Tab. \ref{table:transfer_comparison_all_in_one}; Fig. \ref{fig:bar_padchest}, \ref{fig:line_padchest_natural_fewfullshot}, see also Suppl. material). The transfer improvement is clearly expressed when increasing both model and natural image data pre-training scale across all large X-Ray targets. Interestingly, the effect of scale is still clearly expressed even when conducting inter-domain transfer in a strongly unfavorable setting,  pre-training on narrow domain-specific X-Ray data and transferring to generic natural images (grayed out area in Tab. \ref{table:transfer_comparison_all_in_one}). This indicates that scale can still have beneficial effect even in situations where transfer scenario itself is ill-posed. 

Another remarkable finding arises when further comparing performance of intra-domain medical-medical and inter-domain natural-medical transfer. The largest ResNet-152x4 pre-trained on the largest generic natural ImageNet-21k turns out to be as good or in many cases better than any network pre-trained on the largest available medical domain specific X-Ray superset data when performing full shot transfer to large X-Ray targets (Tab. \ref{table:transfer_comparison_all_in_one}, Figs. \ref{fig:bar_padchest}, \ref{fig:line_padchest_natural_fewfullshot}, \ref{fig:line_padchest_medical_fewfullshot}). The finding indicates that by substantially increasing model and generic source natural image data scale during pre-training, we can obtain models for transfer to medical domain-specific X-Ray images that match or even outperform models pre-trained with large amounts of domain-specific X-Ray data, which may be often not available in practice. The finding also strengthens previously inconclusive evidence for benefits of large-scale pre-training on generic natural images when transferring to medical targets~\cite{Mustafa2021}, providing direct comparison between models pre-trained either on out-of-domain natural images or on in-domain medical X-Ray chest imaging data when transferring to medical image targets. The observation expands the previous evidence that strongly increasing model and data pre-training scale improves transfer also for larger out-of-domain targets - scale of generic ImageNet-21k  (14M samples) is more than order of magnitude larger than scale of the largest medical X-Ray superset we have constructed for this study ($\approx$ 0.87M samples).





For few-shot inter-domain transfer or for transfer on smaller targets, we do not see consistent positive effect of larger pre-training scale (Figs. \ref{fig:bar_padchest}, \ref{fig:line_fewfullshot}). Thus, we again obtain differential picture of scaling benefits depending on transfer scenario. Further scaling up of model and data during pre-training may homogenize this picture and make scaling benefit look more consistent through different conditions, showing improvement for smaller targets and for few-shot transfer. For instance, we could not use substantially larger datasets like JFT-300M or JFT-3B~\cite{Sun2017,Kolesnikov2020,zhai2021scaling} which are proprietary and not available publicly. Additionally, computational budget was here not enough to experiment with networks larger than ResNet-152x4. However, there may be also fundamental limitations prohibiting transfer improvement no matter how large pre-training scale may become that are due to strong incompatibility between source and target domains. There is some evidence from language modeling studies that hints on such fundamental limitations for transfer improvement on target datasets far from source when doing straightforward scaling without further changes in model architecture. For instance, the work by Hendrycks et al.~\cite{Hendrycks2021} finds no improvement of transfer when increasing size of large Transformer networks pre-trained on very large conventional text datasets and fine-tuning those on a specific target dataset (mathematical text tasks of advanced difficulty) far apart from the source.

\textbf{Limitations of the current study.} There are several limitations of the current study that impede more general conclusions about effect of pre-training scale on intra- and inter-domain transfer from the observations made in this study. In the conducted transfer experiments, we made use of a heuristic hyper-parameter selection rule - BiT-HyperRule, as introduced in~\cite{Kolesnikov2020} - that determines pre-training hyper-parameters directly from target datasets on which transfer is to be performed. This rule may be heavily biased towards transfer on natural image datasets, as those were the targets used in the original study. If modifying the rule to take also target domain - natural or medical - into account, the derived hyper-parameters may serve a much better basis for fine-tuning during transfer. In general, performing hyper-parameter tuning for training procedure can strongly boost performance~\cite{Shallue2019}, and this is no different for transfer procedure. Therefore, it cannot be excluded that performing hyper-parameter tuning for each transfer task would alter the effect of larger pre-training scale on transfer. Hyper-parameter tuning would however also impose further cost on transfer that is avoided by employing the hyper-rule. 

We also have not explored other backbone network architectures except the standard ResNet. Although ResNet has proven itself a versatile network architecture for dealing with various vision tasks, it cannot be ruled out that while for instance the inductive bias inherent to its convolutional design is well suited for working on natural image statistics with strong local spatial correlations, it may be less suited for providing good basis for generalization when dealing with other types of image signals. Scaling up ResNet architecture may thus be a viable strategy to improve generalization capability on natural image data, while other, more generic architectures, may be required to benefit from scaling in the same way across more diverse data types. We also did not experiment with larger datasets than ImageNet-21k - as those are still mostly proprietary and were not publicly available, as it is the case for JFT-300M~\cite{Sun2017,Kolesnikov2020} or JFT-3B~\cite{zhai2021scaling}. For medical imaging domain, we could not experiment with increasing data scale substantially, as the amount of openly available X-Ray chest data is currently still limited. Finally, we studied the dependence of transfer improvement on pre-training scale exclusively in supervised classification problem setting. Promising work is also done on pre-training in unsupervised fashion with unlabeled data~\cite{Chen2020a, Xie2020, Zoph2020}, where benefits of scaling up pre-training for transfer may turn out to be as well substantial.


\textbf{Conclusion and outlook.} Here, we presented here evidence that substantially increasing model and data scale in the pre-training provides benefits for both intra- and inter-domain transfer across various target datasets from natural and medical X-Ray image domain. The effect of pre-training scale on transfer performance depends on transfer scenario. Transfer improvement due to larger pre-training scale was found to be substantial in natural-natural or medical-medical, intra-domain transfer scenarios where source and target datasets were closely related, being especially strongly pronounced in the few-shot transfer regime for natural-natural case and concentrated in full-shot scenario in medical-medical case. For natural-medical inter-domain transfer, clear positive effect of larger pre-training scale was found for full shot transfer on large X-Ray targets. On small X-Ray targets and for few-shot transfer regime, no clear inter-domain transfer improvements were observed. Remarkably, the largest ResNet-152x4 network pre-trained on very large generic natural ImageNet-21k matched or even outperformed networks pre-trained on largest medical domain-specific X-Ray superset data combined for this study when performing full shot transfer to large X-Ray targets. This is relevant for the practice, as large amount of medical domain-specific data is often not available for pre-training. Here we show that high quality models for large X-Ray targets can be also obtained by substantially increasing pre-training model and generic natural image source data scale instead, obliterating need for large domain-specific data.

The study offers different follow-up directions. One of these are experiments with larger scale both for network and data size, for instance going beyond ImageNet-21k, or combining in the pre-training different source datasets that may contain both natural and medical images. This may also include experiments with scaling up other architectures than ResNet. Another direction to study effect of scale is to employ various unsupervised learning strategies for pre-training~\cite{Chen2020a, Xie2020,Zoph2020} instead of supervised learning. Yet another fruitful path is to provide a measure of source and target domain similarity and to experiment with more than two distinct domains, varying systematically relatedness between different source and target data. First steps in this direction for language modeling was already undertaken in~\cite{Hernandez2021}. Following these directions would pave the path towards scaling laws for transfer in the image domain, taking into account different pre-training regimes and affinity between source and target domains, to enable systematic prediction of transfer performance and improvement due to increase of pre-training scale.

\section*{Broader and Social Impact}
\label{sec:impact}

Our work aims on advancing transfer learning, which can make learning algorithms perform better and more efficient by re-using models already pre-trained on various tasks and therefore requiring less compute and data to learn solutions for other relevant tasks. The approach to improve transfer learning by increasing scale of the pre-training is generic and has impact far beyond vision domain, for instance in language modeling, and is not bound to any specific application. As any generic method, it can be therefore applied to enhance technologies for sensitive applications, for instance in health domain or in public surveillance, that may have both strong positive and negative social impact, depending on policies introduced on their usage. Special care should be taken about applications in clinical domain where further development of diagnostic tools based on data driven machine learning should be accompanied by a broad panel of experts from corresponding domains. The method depends on computationally heavy large-scale pre-training that is energy demanding on the one hand. On the other hand, it contains a promise to pay off the energy budget put into training by obtaining generic models that can be very efficiently adapted to a large range of problems via transfer, saving computational and energy costs that would otherwise incur for their solution from scratch.

\begin{ack}
We would like to express gratitude to all the people who are working on making code, models and data publicly available, advancing community based research and making research more reproducible. Special thanks go to creators and maintainers of open available X-Ray medical imaging datasets that also enabled our research, some of those gathered under difficult circumstances of the COVID-19 pandemics. The authors gratefully acknowledge the Gauss Centre for Supercomputing e.V. (www.gauss-centre.eu) for funding this work by providing computing time through the John von Neumann Institute for Computing (NIC) on the GCS Supercomputers JUWELS, JUWELS Booster at Jülich Supercomputing Centre (JSC). We also acknowledge computing resources from the Helmholtz Data Federation and further computing time provided on supercomputer JUSUF in frame of offer for epidemiology research on COVID-19 by JSC.
\end{ack}

\bibliographystyle{unsrt}
\bibliography{effect_transfer_paper_2021}

\clearpage

\begin{appendix}

\begin{center}
{\Large\bf Supplementary: Effect of Pre-Training Scale on Intra- and Inter-Domain Full and Few-Shot Transfer Learning for Natural and Medical X-Ray Chest Images}
\end{center}


\section{Distributed Training}
\label{appendix:distributed_training}


\subsection{JUWELS Booster Supercomputer}
\label{appendix:booster}

Installed in November 2020, JUWELS Booster~\cite{JUWELSBooster2020} features \num{936} compute nodes that host four NVIDIA A100 GPUs each, providing \num{3744}~GPUs in total. The installed A100 Tensor Core GPUs~(\SI{40}{\giga\byte}) provide \SI{19.5}{\tera\floppersec} of $\text{FP64}_\text{TC}$ computing performance each. The GPUs are hosted by AMD EPYC 7402 CPUs with $2\times 24$~cores (SMT-2) per node, clocked with \SI{2.8}{\giga\Hz}. Each node is diskless and is equipped with \SI{512}{\giga\byte} of RAM. The network of JUWELS Booster is based on Mellanox HDR200 InfiniBand, with four Mellanox ConnectX~6 devices per node, each providing \SI{200}{\giga\bit\per\second} bandwidth per direction.

The NVIDIA A100 GPUs installed into JUWELS Booster reach peak efficiency of \SI{48,75}{\giga\floppersec\per\watt} when utilizing the $\text{FP64}$ Tensor Cores. This makes JUWELS Booster rank highest in the Green500 list of November 2020 as the most energy efficient supercomputer among the first 100 machines of the Top500 list with \SI{25}{\giga\floppersec\per\watt}.


\subsection{Scaling and training time}
\label{appendix:scaling}

Here, we report scaling behavior during large-scale pre-training for ResNet networks we used in the experiments.

\begin{figure*}[ht]
\begin{subfigure}{.5\textwidth}
  \centering
  \includegraphics[width=\textwidth]{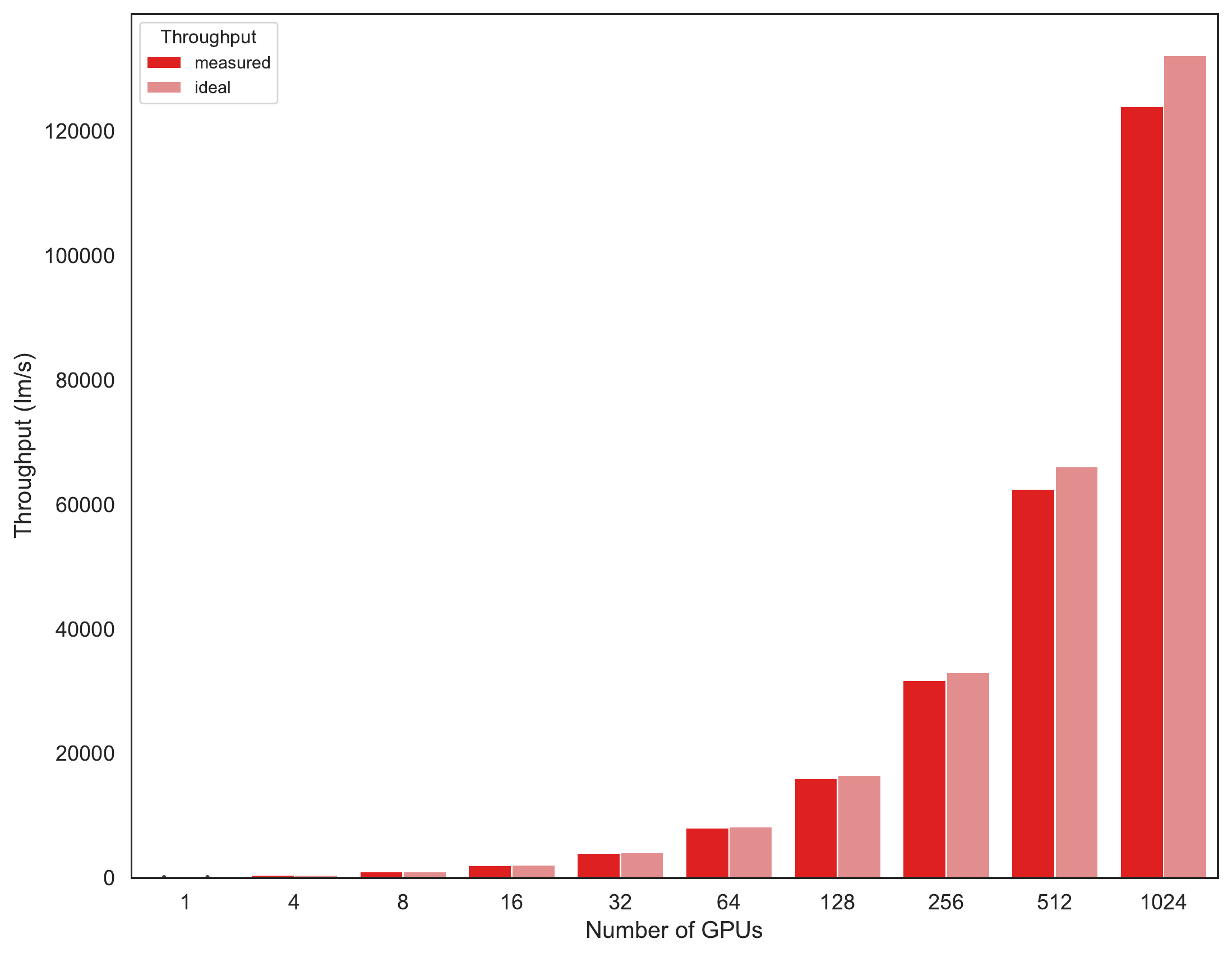}
  \caption{Throughput in Im/s}
  \label{fig:throughput_R152}
\end{subfigure}
\begin{subfigure}{.5\textwidth}
  \centering
  \includegraphics[width=\textwidth]{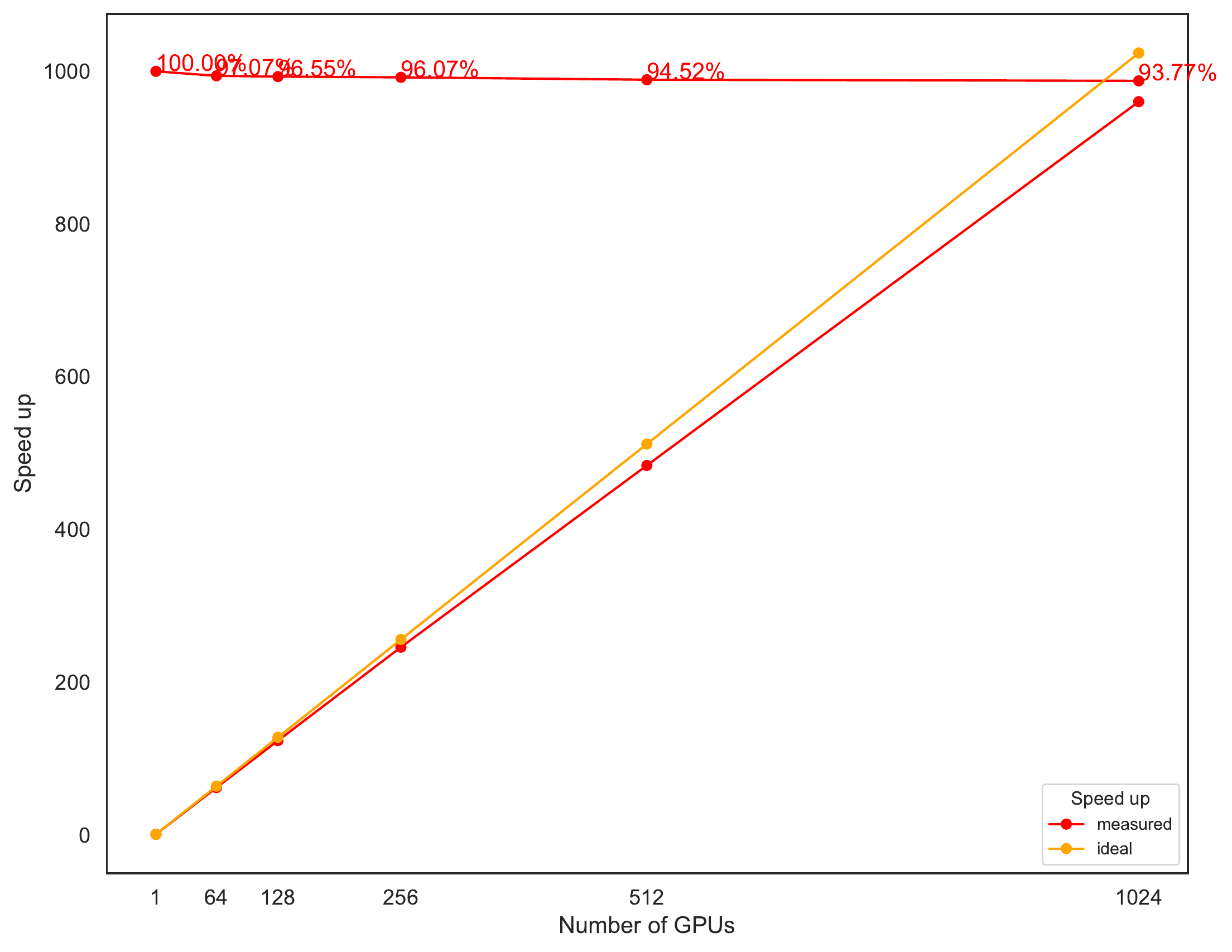}
  \caption{Scaling efficiency}
  \label{fig:scaling_efficiency_R152}
\end{subfigure}
\vspace*{-0.2cm}
\caption{Distributed training for R152x4, scaling behavior on JUWELS Booster using A100 GPUs.}
\label{fig:scaling_R152}
\end{figure*}

We performed scaling experiments to assess the scalability of data parallel training distributed across many GPUs on multiple nodes using Horovod.
The efficiency in Figure \ref{fig:scaling_efficiency_R152} (upper part of the figure with percentages) is computed using the following formula: $E(N) = 100 \times \frac{T(N)}{N \times T(1)}$. $T(N)$ is the total measured throughput in Im/s for $N$ GPUs. The best achievable efficiency, when scaling is perfect, is 100\%.

We also provide the raw throughput (Im/s) numbers in Figure \ref{fig:throughput_R152} and Tab. \ref{tab:juwelsbooster_scaling}. 
On 1024 GPUs, we achieve an efficiency of $\approx 93.7\%$ with single precision (FP32). To make sure distributed training is stable, we check the end accuracy of full training for each number of GPUs to reassure we reach target accuracy acceptable for standard ImageNet-1k Top-1 and Top-5 results.

Achieved scaling on JUWELS Booster allows to perform full pre-training on ImageNet-21k with large R152x4 in about 81 hours using $256$ GPUs. For small R50x1, full training needs about 13.5 hours to finish using $128$ GPUs.

\begin{table}[tbh!]
\footnotesize
	\caption{\label{imgs_throughput_scaling_imagenet} Scaling behavior in Im/s of ImageNet-1k training using ResNet-152x4 architecture from \cite{Kolesnikov2020} with batch size 128. For each GPU, one MPI process is assigned. Computations were done on {up to 256 nodes on JUWELS Booster}. Throughput performance during training is reported for single precision mode (FP32). The corresponding speedup is provided relative to reference training with 1 GPU. Note that the measured Im/s throughput includes I/O.}
	\begin{minipage}{\textwidth}
		\begin{center} \small
			\setlength{\tabcolsep}{10pt}
			\renewcommand{\arraystretch}{1}
			\begin{tabular}{lrr}
              \#GPUs  & Im/s & speedup \\
                        \hline	
                1    & \phantom1      129.14 & 1.00 \\
                4    & \phantom1      508.00 & 3.93 \\
                8    & \phantom1      1009.30 & 7.82 \\
                16    & \phantom1      2023.78 & 15.67 \\
                32    & \phantom1      4029.69 & 31.21  \\
                64    & \phantom1      8022.31 & 62.12  \\
                128    & \phantom1      15959.86 & 123.59  \\
                256    & \phantom1      31758.59 & 245.93  \\
                512    & \phantom1      62496.35 & 483.96  \\
                1024    & \phantom1      124003.59 & 960.26  \\
			\end{tabular}
		\end{center}	
	\end{minipage}%
	\label{tab:juwelsbooster_scaling}
\end{table}


\section{Additional details on experimental results}
\label{appendix:results}

All datasets employed in our experiments are publicly available and can be obtained following links in the Tab. \ref{tab:datasets_list}



\subsection{Further transfer results}

Here, we present more detailed results of transfer experiments described in the main document. For medical X-Ray targets, we provide tables reporting transfer performance (Tabs. \ref{table:transfer_medicalsource_shorter}, \ref{table:transfer_medicalsource_mimic}, \ref{table:transfer_medicalsource_chexpert}, \ref{table:transfer_medicalsource_padchest}, \ref{table:transfer_medicalsource_nih}) listing each source X-Ray dataset and supersets used for pre-training, as outlined in the experiments description in the main document. 


\begin{figure*}[ht]
\begin{subfigure}{.5\textwidth}
  \centering
  \includegraphics[width=\linewidth]{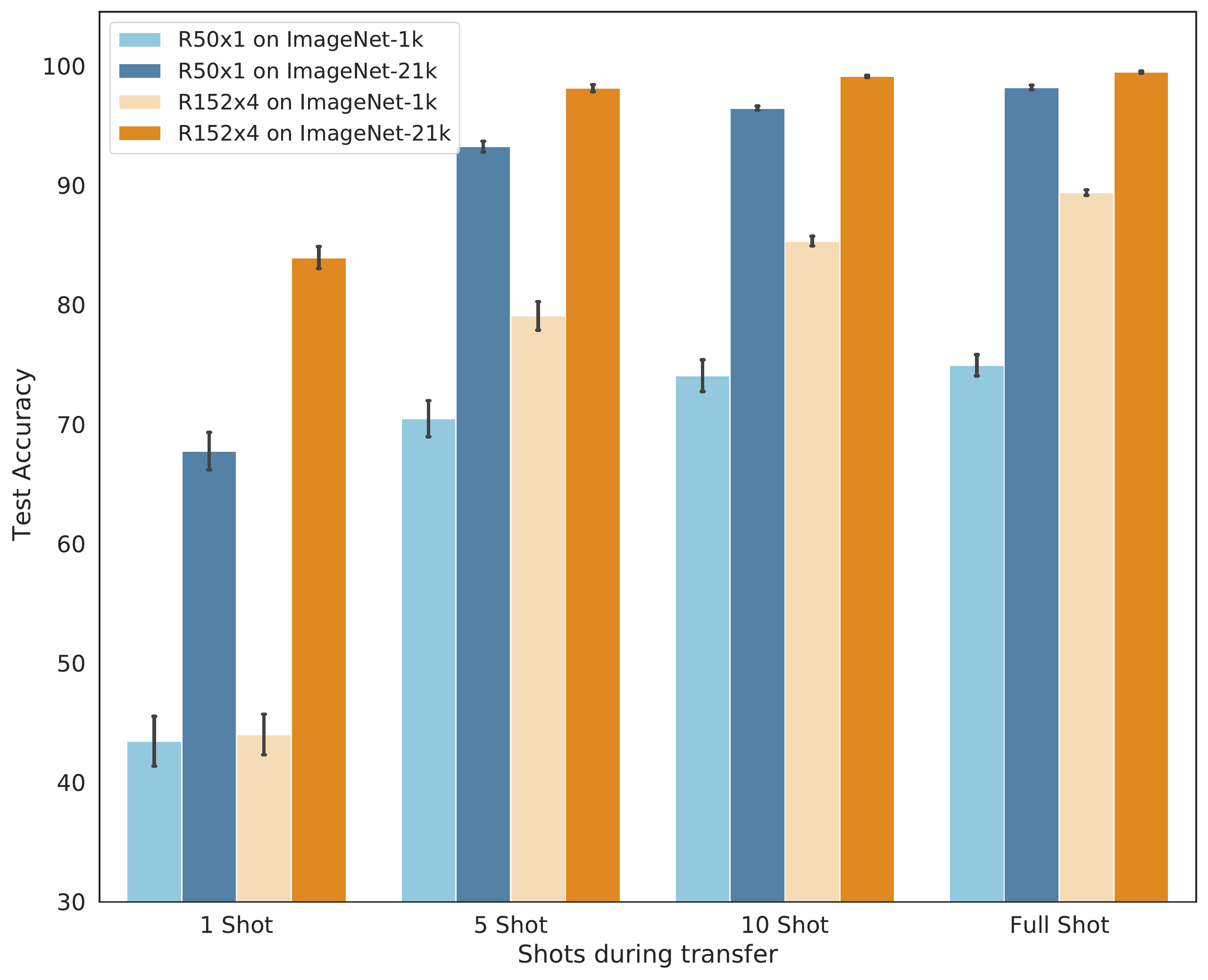}
  \caption{Flowers-102}
  \label{fig:bar_flowers}
\end{subfigure}
\begin{subfigure}{.5\textwidth}
  \centering
  \includegraphics[width=\linewidth]{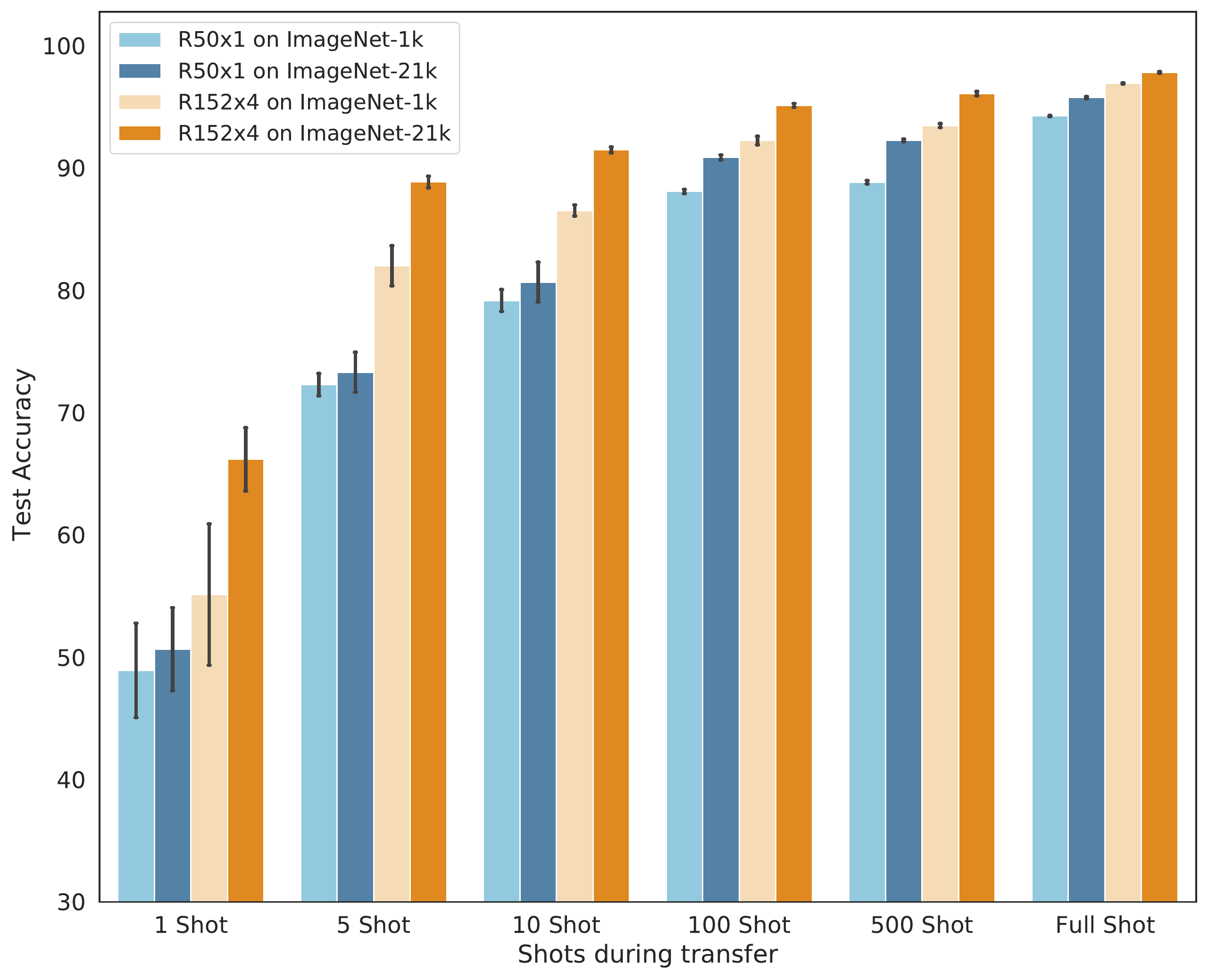}
  \caption{CIFAR-10}
  \label{fig:bar_cifar10}
\end{subfigure}
\vspace*{-0.2cm}
\caption{
Few-shot and full shot transfer performance on target datasets when varying model size and dataset size in pre-training. Transfer improvement due to model and source data size is evident, especially strongly pronounced in few-shot regime.
}
\end{figure*}

\begin{figure*}[ht]
  \centering
  \includegraphics[width=0.8\linewidth]{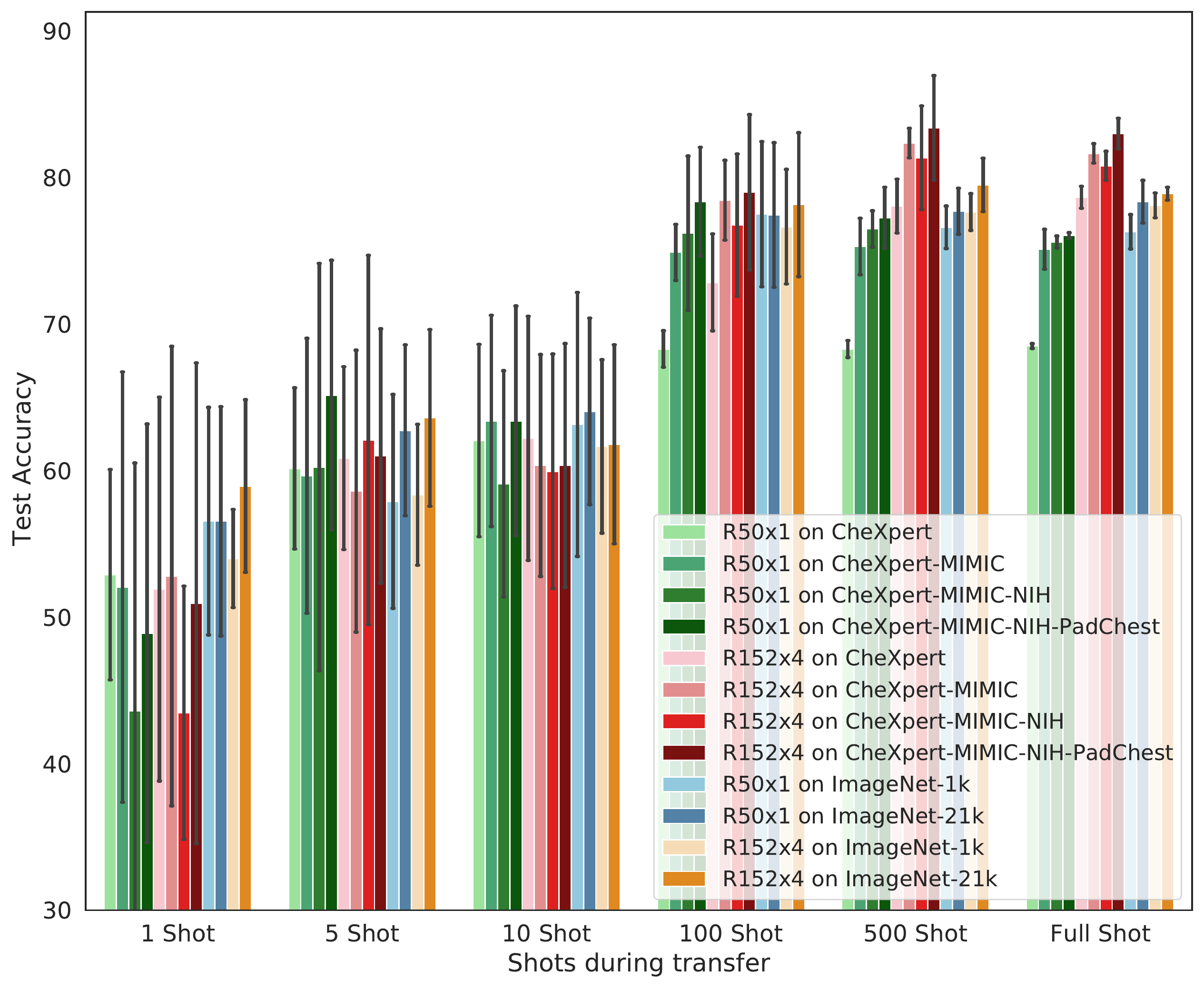}
\vspace*{-0.2cm}
\caption{Few-shot and full shot transfer performance on COVIDx dataset when varying model, source data size and domain in pre-training. In the full shot transfer, improvement due to model and data scale is evident when pre-training on X-Ray chest imaging source data. In few-shot regime, no transfer improvement due to larger model or data size is observed.}
\label{fig:bars_covidx}
\end{figure*}

\begin{figure*}[ht]
  \centering
  \includegraphics[width=0.8\linewidth]{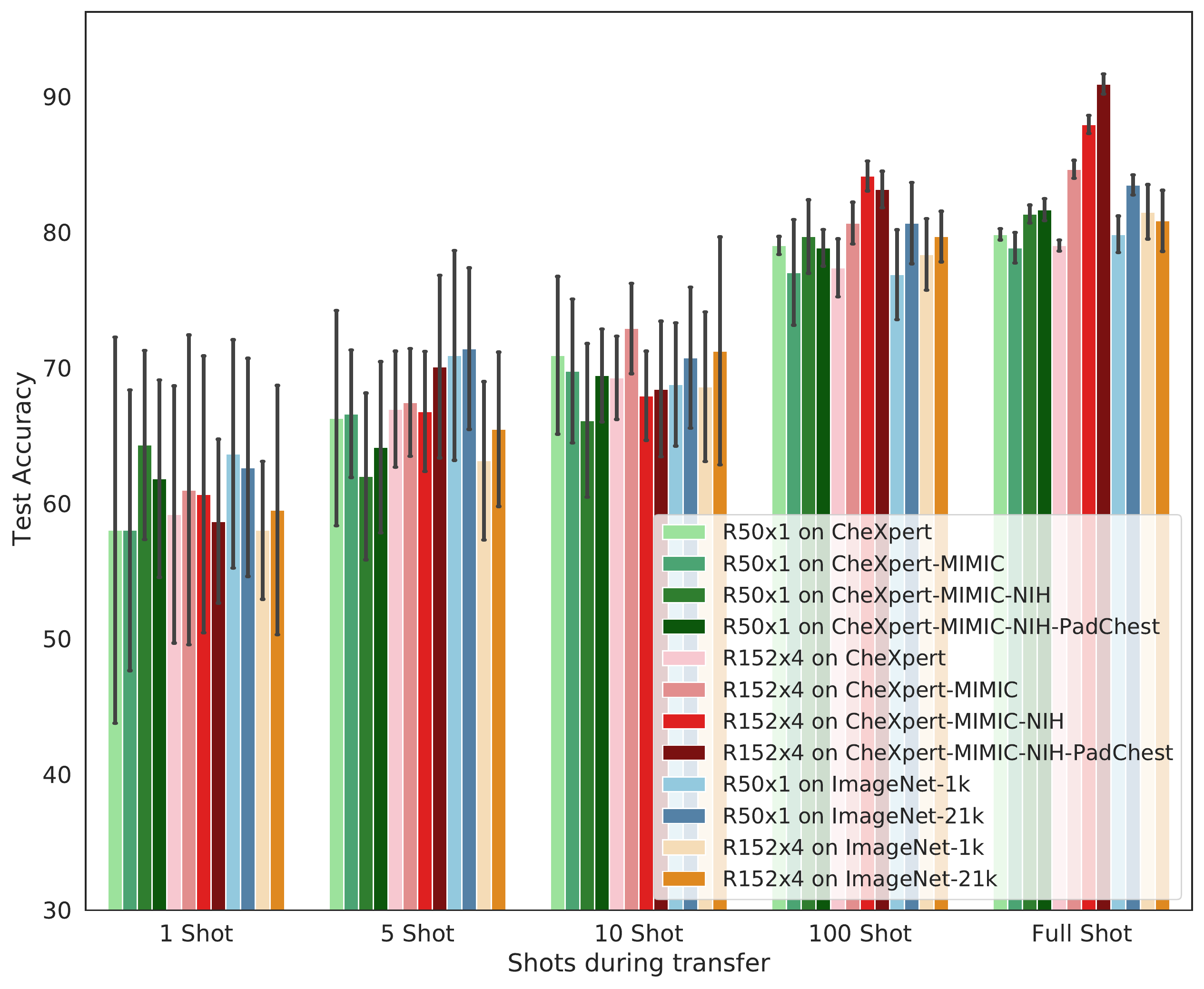}
\vspace*{-0.2cm}
\caption{Few-shot and full shot transfer performance on Tuberculosis dataset when varying model, source data size and domain in pre-training. In the full shot transfer, improvement due to model and data scale is evident when pre-training on X-Ray chest imaging source data. In few-shot regime, no transfer improvement due to larger model or data size is observed.}
\label{fig:bars_tuberculosis}
\end{figure*}

\begin{figure*}[ht]
\begin{subfigure}{.5\textwidth}
  \centering
  \includegraphics[width=\linewidth]{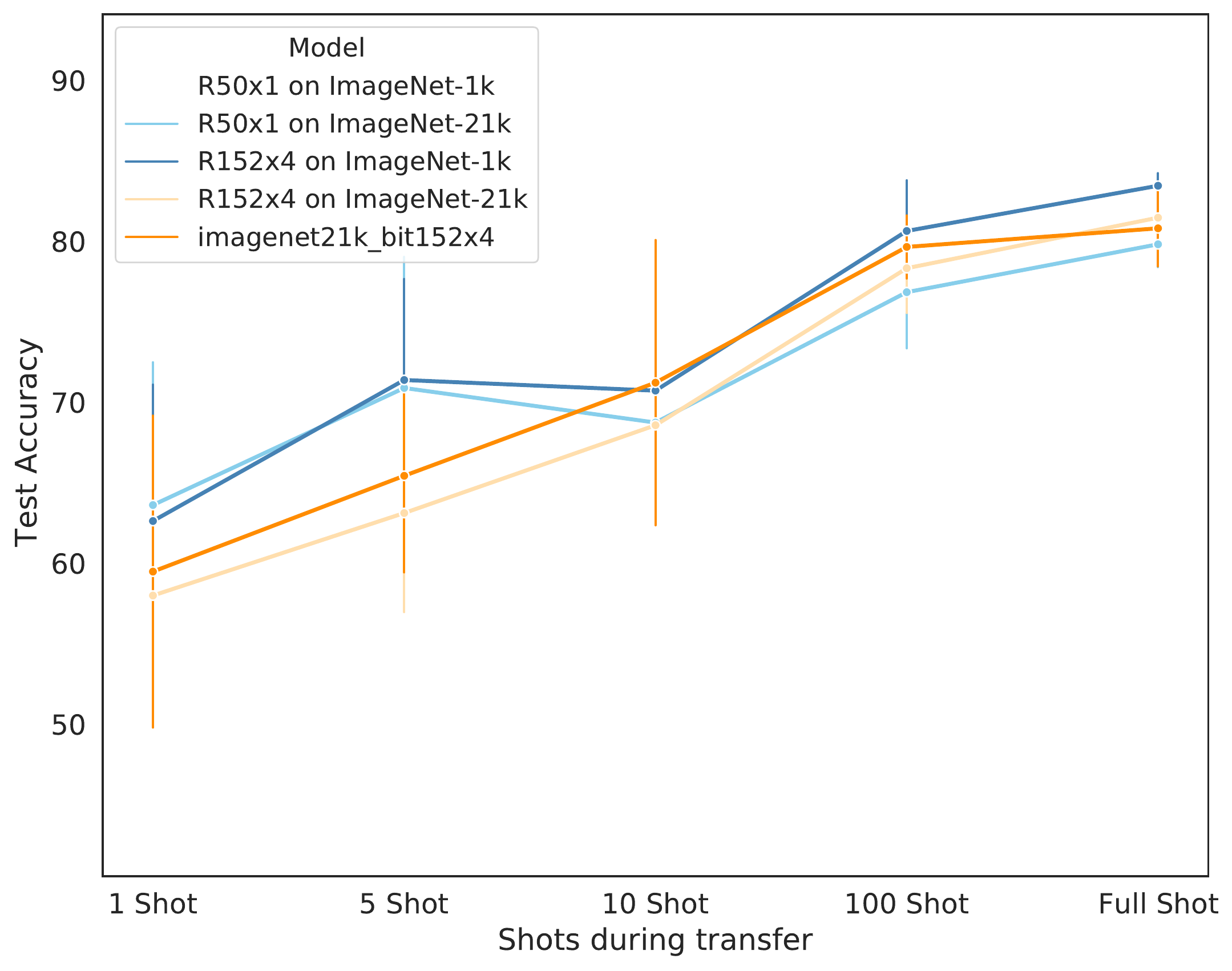}
  \caption{Tuberculosis, natural sources}
  \label{fig:appendix_line_tuberculosis_natural_fewfullshot}
\end{subfigure}
\begin{subfigure}{.5\textwidth}
  \centering
  \includegraphics[width=\linewidth]{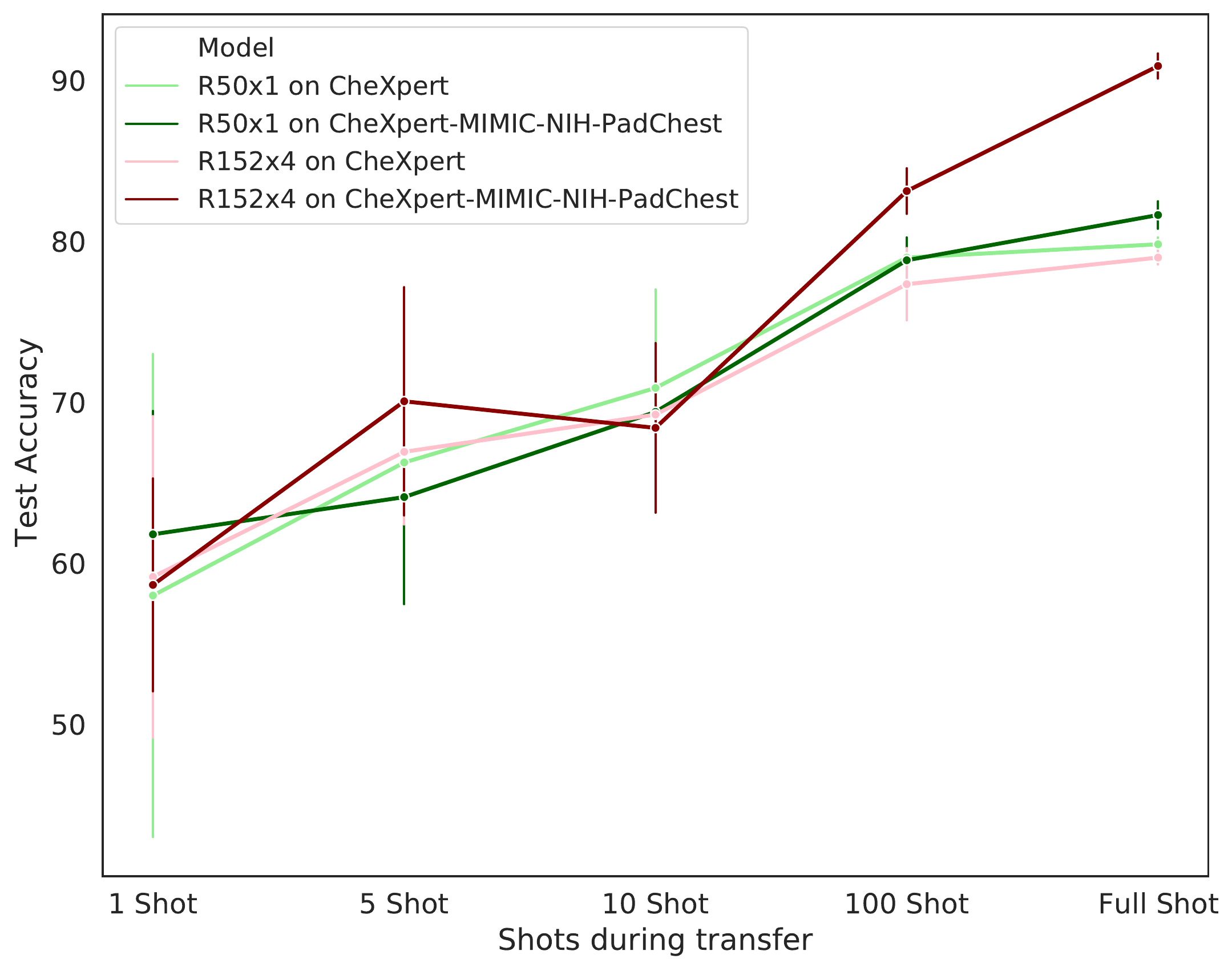}
  \caption{Tuberculosis, medical sources}
  \label{fig:appendix_line_tuberculosis_medical_fewfullshot}
\end{subfigure}

\caption{
Few-shot and full shot transfer performance on Tuberculosis dataset when pre-training with different model sizes on different sources (natural or medical datasets) of various sizes. In natural-medical scenario \textbf{(a)}, no transfer improvement due to model or data scale is evident. In medical-medical scenario \textbf{(b)}, larger model and data size lead to transfer improvement in full shot regime, without benefits in few-shot mode.}
\label{fig:appendix_tuberculosis_line_fewfullshot}
\end{figure*}


\begin{figure*}[ht]
\begin{subfigure}{.33\textwidth}
  \centering
  \includegraphics[width=\linewidth]{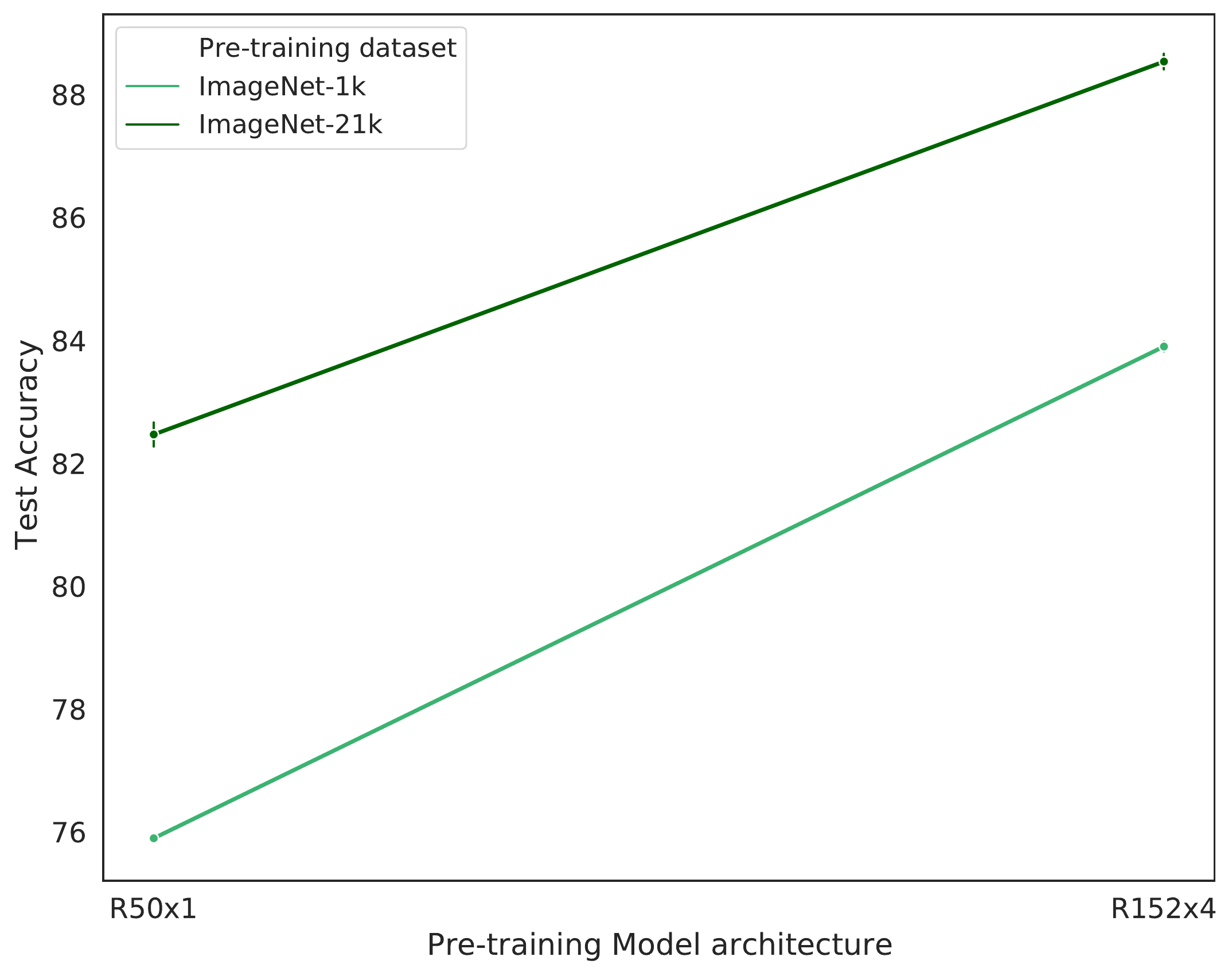}
  \caption{CIFAR-100}
  \label{fig:fullshot_cifar100}
\end{subfigure}
\begin{subfigure}{.33\textwidth}
  \centering
  \includegraphics[width=\linewidth]{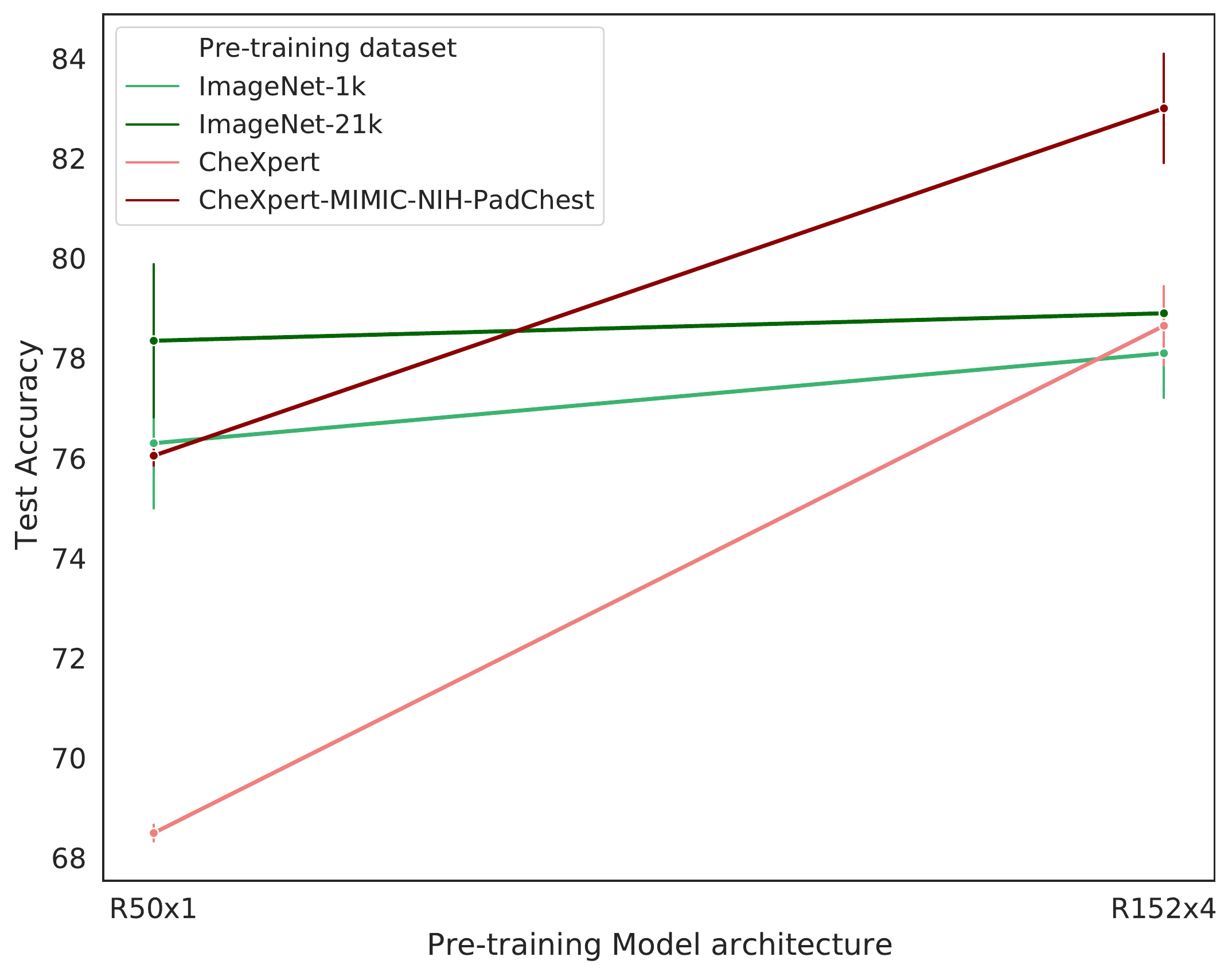}
  \caption{COVIDx}
  \label{fig:fullshot_covidx}
\end{subfigure}
\begin{subfigure}{.33\textwidth}
  \centering
  \includegraphics[width=\linewidth]{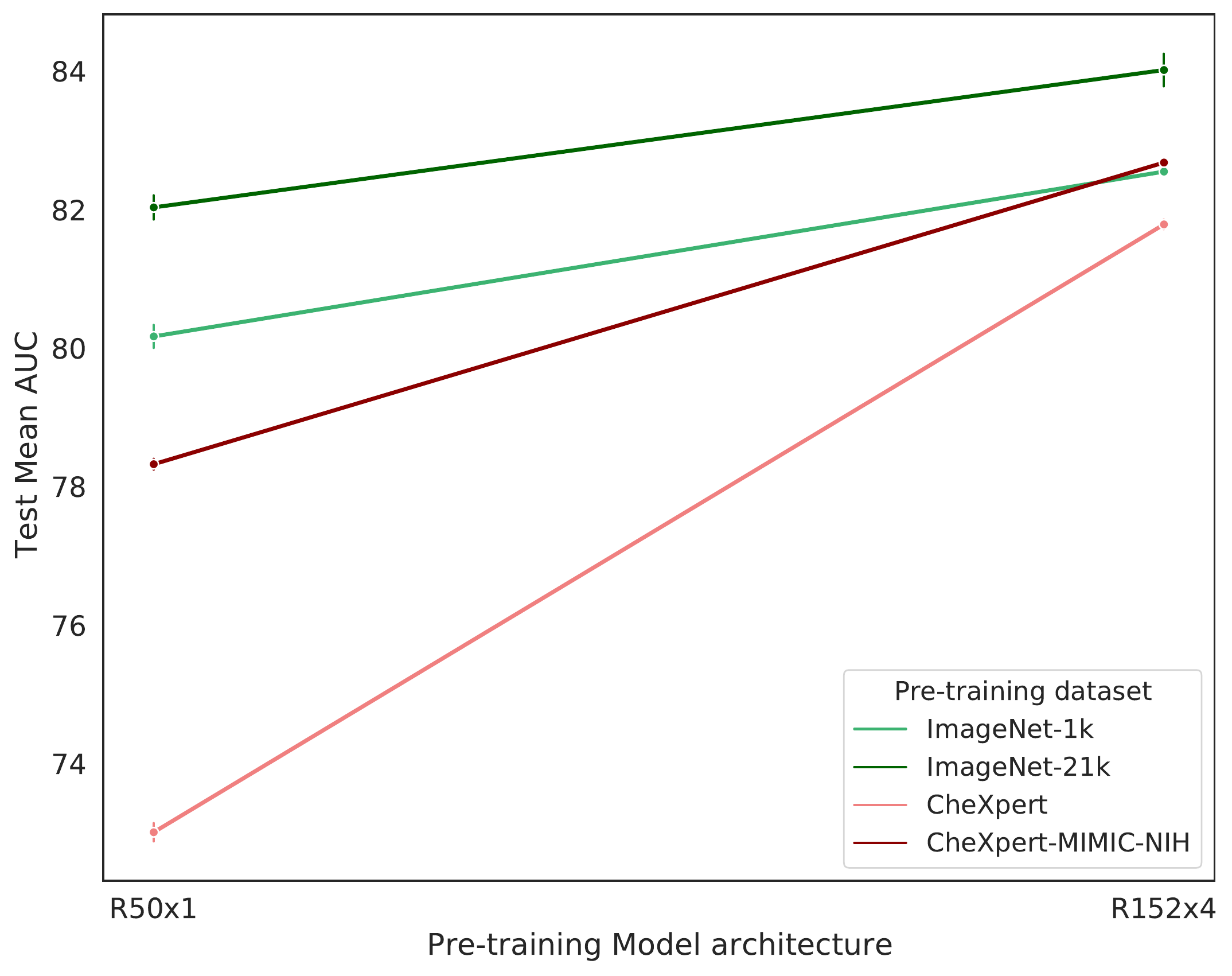}
  \caption{PadChest-Cl}
  \label{fig:fullshot_padchest}
\end{subfigure}
\caption{
Full shot transfer performance on target datasets when varying model and source data size, taking the smallest and largest pre-training datasets available for each domain.}
\label{fig:line_fullshot}
\end{figure*}

\begin{table*}[t!]
\caption{\textbf{Intra- and inter-domain transfer using natural ImageNet-1k and ImageNet-21k for pre-training with different sized ResNets} (1) - Top-1 Acc [$\%$] metric; (2) - mean AUC metric. Bold indicates best transfer performance for a fixed network size and pinpoints the effect of data scale on transfer. Italics indicates transfer performance with no significant difference between data scale. \textcolor{red}{Red} indicates best overall performance for a given target. Clear transfer improvement emerges for natural-natural scenario due to both model and data scale. For natural-medical scenario the positive effect of larger scale is consistently given for larger targets, but not for smaller ones. For instance, for very small Tuberculosis target, larger data scale improves transfer for small ResNet-50x1, while larger model scale does not lead to any transfer improvement.}
\centering
\begin{small}
\begin{tabular}{|c||c||cc||cc||}
\hline
\multirow{2}{*}{\begin{tabular}[c]{@{}c@{}}Target\\ Domain\end{tabular}}  & \multirow{2}{*}{Dataset} & \multicolumn{2}{c||}{ResNet-50x1} & \multicolumn{2}{c||}{ResNet-152x4}  \\ 
\cline{3-6}
                &                          & 1K   & 21K     & 1K   & 21K      \\  
\hhline{=::=::==::==:|}
\multirow{4}{*}{\begin{tabular}[c]{@{}c@{}}Natural \end{tabular}} 
 & CIFAR-10\textsuperscript{(1)}      & 94.26 $\pm$ 0.05 & \textbf{95.78 $\pm$ 0.09} & 96.93 $\pm$ 0.05 & \color{red} \textbf{97.82 $\pm$ 0.07}  \\ 
 & CIFAR-100\textsuperscript{(1)}     & 75.90 $\pm$ 0.05 & \textbf{82.47 $\pm$ 0.21} & 83.90 $\pm$ 0.09 & \color{red} \textbf{88.54 $\pm$ 0.14}  \\ 
 & Flowers-102 \textsuperscript{(1)}  & 74.94 $\pm$ 0.99 & \textbf{98.21 $\pm$ 0.22} & 89.41 $\pm$ 0.25 & \color{red} \textbf{99.49 $\pm$ 0.08}  \\ 
& Pets \textsuperscript{(1)}  & 85.21 $\pm$ 0.58 & \textbf{87.23 $\pm$ 0.18} & \color{red} \textit{93.32 $\pm$ 0.30} & \color{red} \textit{93.21 $\pm$ 0.14}  \\
 \hhline{=::=::==::==:|}
\multirow{7}{*}{\begin{tabular}[c]{@{}c@{}}Medical\end{tabular}}  
& Tuberculosis\textsuperscript{(1)}   & 79.83 $\pm$ 1.50 & \color{red} \textit{ \textbf{83.47 $\pm$ 0.83}} & \color{red} \textit{81.49 $\pm$ 2.23} & \color{red} \textit{80.83 $\pm$ 2.51}  \\
& COVIDx\textsuperscript{(1)}         & \textit{76.30 $\pm$ 1.30} & \textit{78.35 $\pm$ 1.63} & \textit{78.10 $\pm$ 0.95} & \textit{78.90 $\pm$ 0.49}  \\
& NIH \textsuperscript{(2)}   & 75.53 $\pm$ 0.47 & \textbf{81.02 $\pm$ 0.57} & 79.82 $\pm$ 0.38 & \color{red} \textbf{82.80 $\pm$ 0.41}  \\
& PadChest-Cl\textsuperscript{(2)}       & 80.17 $\pm$ 0.17 & \textbf{82.03 $\pm$ 0.17} & 82.55 $\pm$ 0.05 & \color{red} \textbf{84.02 $\pm$ 0.24}  \\ 
& PadChest \textsuperscript{(2)}   & 76.72 $\pm$ 0.27 & \textbf{80.99 $\pm$ 0.22} & 79.59 $\pm$ 0.17 & \color{red} \textbf{83.94 $\pm$ 0.19}  \\
& CheXpert \textsuperscript{(2)}   & 84.83 $\pm$ 0.14 & \textbf{86.60 $\pm$ 0.14} & 86.82 $\pm$ 0.06 & \color{red} \textbf{87.77 $\pm$ 0.07}  \\ 
& MIMIC CXR\textsuperscript{(2)}   & 85.41 $\pm$ 0.10 & \textbf{86.82 $\pm$ 0.10} & 86.85 $\pm$ 0.06 & \color{red} \textbf{87.79 $\pm$ 0.13}  \\ 

\hline
\end{tabular}
\end{small}
\label{table:transfer_naturalsource_comparison}
\end{table*}

\begin{table*}[t!]
\caption{\textbf{Intra-domain transfer using different sized medical X-Ray source data for pre-training with different sized ResNets} (1) - Top-1 Acc [$\%$] metric; (2) - mean AUC metric. "+" indicates addition into a successively larger source superset. Clear transfer improvement is evident due to larger model and data scale across different targets.}
\centering
\begin{scriptsize}
\begin{tabular}{|c||cccc||cccc||}
\hline
\multirow{2}{*}{Target} & \multicolumn{4}{c||}{ResNet-50x1} & \multicolumn{4}{c||}{ResNet-152x4}  \\ 
\cline{2-9}
                                      & CheXpert   & +MIMIC  & + NIH & +PadChest & CheXpert  & +MIMIC  & +NIH & +PadChest      \\  
\hhline{=::====::====:|}  
PadChest-Cl\textsuperscript{(2)}       & 73.01 $\pm$ 0.13 & \textit{78.44 $\pm$ 0.04} & \textit{78.33 $\pm$ 0.08} & --- & 81.79 $\pm$ 0.07 & \color{red} \textbf{83.14 $\pm$ 0.04} & 82.68 $\pm$ 0.05 & ---  \\ 
COVIDx\textsuperscript{(1)}         & 68.50 $\pm$ 0.18 & 75.10 $\pm$ 1.52 & 75.60 $\pm$ 0.45 & \textbf{76.05 $\pm$ 0.21} & 78.65 $\pm$ 0.84 & 81.65 $\pm$ 0.74 & 80.80 $\pm$ 1.10 & \color{red} \textbf{83.00 $\pm$ 1.16}  \\
Tuberculosis\textsuperscript{(1)}   & 79.83 $\pm$ 0.45 & 78.84 $\pm$ 1.25 & \textit{81.32 $\pm$ 0.74} & \textit{81.65 $\pm$ 0.91} & 79.01 $\pm$ 0.45 & 84.63 $\pm$ 0.74 & 87.93 $\pm$ 0.74 & \color{red} \textbf{90.91 $\pm$ 0.83}  \\
\hline
\end{tabular}
\end{scriptsize}
\label{table:transfer_medicalsource_shorter}
\end{table*}

\begin{table*}[t!]
\caption{\textbf{Intra-domain transfer using different sized medical X-Ray source data for pre-training with different sized ResNets, target MIMIC-CXR} Mean AUC metric. "+" indicates addition into a successively larger source superset. Clear transfer improvement is evident by scaling the model size. Using a superset containing CheXpert and PadChest improves the results, but adding NIH does not or does very little, this could be explained by the fact that NIH is the smallest dataset among the medical pre-training datasets, and a larger increase in the superset would be needed to substantially improve the transfer results, as it has been observed in transfer results that were obtained using models pre-trained on much larger natural data.}
\centering
\begin{scriptsize}
\begin{tabular}{|c||ccc||ccc||}
\hline
\multirow{2}{*}{Target} & \multicolumn{3}{c||}{ResNet-50x1} & \multicolumn{3}{c||}{ResNet-152x4}  \\ 
                                      & CheXpert   & +PadChest  & + NIH  & CheXpert  & +PadChest  & +NIH  \\  
\hhline{=::===::===:|}  
MIMIC CXR       & 84.17 $\pm$ 0.03 & 86.19 $\pm$ 0.03 & \textbf{86.38 $\pm$ 0.03} & 87.63 $\pm$ 0.04 & \color{red} \textbf{88.13 $\pm$ 0.03} & 88.00 $\pm$ 0.03  \\ 
\hline
\end{tabular}
\end{scriptsize}
\label{table:transfer_medicalsource_mimic}
\end{table*}

\begin{table*}[t!]
\caption{\textbf{Intra-domain transfer using different sized medical X-Ray source data for pre-training with different sized ResNets, target CheXpert} Mean AUC metric. "+" indicates addition into a successively larger source superset. Clear transfer improvement is evident by scaling the model size. Using a superset containing PadChest and MIMIC CXR improves the results, adding NIH does not lead to further improvement. This could be explained by the fact that NIH is the smallest dataset among the medical pre-training datasets, and a larger increase in the superset would be needed to substantially improve the transfer results, as it has been observed in transfer results that were obtained using models pre-trained on much larger natural data.}
\centering
\begin{scriptsize}
\begin{tabular}{|c||ccc||ccc||}
\hline
\multirow{2}{*}{Target} & \multicolumn{3}{c||}{ResNet-50x1} & \multicolumn{3}{c||}{ResNet-152x4}  \\ 
                                      & PadChest   & +MIMIC  & + NIH  & PadChest  & +MIMIC  & +NIH  \\  
\hhline{=::===::===:|}  
CheXpert       & 82.10 $\pm$ 0.07 & \textit{86.56 $\pm$ 0.08} & \textit{86.66 $\pm$ 0.05} & 84.92 $\pm$ 0.07 & \color{red} \textbf{88.03 $\pm$ 0.03} & 87.82 $\pm$ 0.03  \\ 
\hline
\end{tabular}
\end{scriptsize}
\label{table:transfer_medicalsource_chexpert}
\end{table*}

\begin{table*}[t!]
\caption{\textbf{Intra-domain transfer using different sized medical X-Ray source data for pre-training with different sized ResNets, target PadChest} Mean AUC metric. "+" indicates addition into a successively larger source superset. Clear transfer improvement is evident by scaling the model size. Improvement by increasing data size is not evident and only happens using the small R50x1 model and a superset containing CheXpert and MIMIC, adding NIH (which is smaller compared to CheXpert and MIMIC) the superset does not help further. This indicates that larger increase in the superset may be necessary to further improve the transfer results, as it has been observed when using models pre-trained on much larger natural data.}
\centering
\begin{scriptsize}
\begin{tabular}{|c||ccc||ccc||}
\hline
\multirow{2}{*}{Target} & \multicolumn{3}{c||}{ResNet-50x1} & \multicolumn{3}{c||}{ResNet-152x4}  \\ 
                                      & CheXpert   & +MIMIC  & + NIH  & CheXpert  & +MIMIC  & +NIH  \\  
\hhline{=::===::===:|}  
PadChest       & 68.06 $\pm$ 0.24 & \textbf{70.07 $\pm$ 0.49} & 68.14 $\pm$ 0.21 & \color{red} \textit{75.91 $\pm$ 0.12} & \color{red} \textit{75.81 $\pm$ 0.07} & 75.23 $\pm$ 0.17  \\ 
\hline
\end{tabular}
\end{scriptsize}
\label{table:transfer_medicalsource_padchest}
\end{table*}

\begin{table*}[t!]
\caption{\textbf{Intra-domain transfer using different sized medical X-Ray source data for pre-training with different sized ResNets, target NIH} (Mean AUC metric). "+" indicates addition into a successively larger source superset.  Clear transfer improvement is evident by scaling the model size. We also observe transfer improvement by scaling data size, however the improvement seems to flatten. Since the transfer results using pre-trained models on larger natural data show a better performance, this indicates that a larger superset scale may be necessary to further improve transfer.}
\centering
\begin{scriptsize}
\begin{tabular}{|c||ccc||ccc||}
\hline
\multirow{2}{*}{Target} & \multicolumn{3}{c||}{ResNet-50x1} & \multicolumn{3}{c||}{ResNet-152x4}  \\ 
                                      & CheXpert   & +PadChest  & + MIMIC  & CheXpert  & +PadChest  & +MIMIC  \\  
\hhline{=::===::===:|}  
NIH      & 70.11 $\pm$ 0.15 & \textit{73.37 $\pm$ 0.38} & \textit{74.21 $\pm$ 0.57} & 77.95 $\pm$ 0.13 & 78.16 $\pm$ 0.13 & \color{red} \textbf{78.95 $\pm$ 0.13}  \\ 
\hline
\end{tabular}
\end{scriptsize}
\label{table:transfer_medicalsource_nih}
\end{table*}

\clearpage
\section{Code and Data availability}
\label{appendix:code}

Repository containing code used for running experiments and producing figures in this study can be found at \url{https://github.com/SLAMPAI/large-scale-pretraining-transfer}. All datasets used in the study are openly available and are listed together with references to the original work in the Table \ref{tab:datasets_list}.  Further details on the usage of the datasets in the conducted experiments are also provided in the linked repository.


\end{appendix}

\end{document}